\documentclass[]{memtensor}

\usepackage[toc,page,header]{appendix}
\usepackage[utf8]{inputenc}
\usepackage[T1]{fontenc}
\usepackage{hyperref}
\usepackage{url}
\usepackage{booktabs}
\usepackage{amsfonts}
\usepackage{nicefrac}
\usepackage{microtype}
\usepackage{amsmath}
\usepackage{amssymb}
\usepackage{etoolbox}
\usepackage{minitoc}
\usepackage[table]{xcolor}
\usepackage{tablefootnote}
\usepackage{threeparttable}
\usepackage{tabularx}
\usepackage{enumitem}
\usepackage{placeins}
\usepackage{fontawesome5}
\usepackage{newunicodechar}
\newcommand{\corrsym}{\raisebox{-0.35ex}{\includegraphics[height=0.52em]{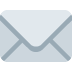}}}
\newunicodechar{✉}{\corrsym}
\newcommand{\iconcor}{\mbox{\raisebox{-0.15ex}{\scriptsize\faIcon[regular]{envelope}}}}
\newcommand{\icongh}{\mbox{\raisebox{-0.15ex}{\scriptsize\faIcon[brands]{github}}}}
\newcommand{\iconhf}{\mbox{\raisebox{-0.15ex}{\includegraphics[height=0.72em]{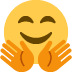}}}}
\newcommand{\iconlegend}{\mbox{\raisebox{-0.15ex}{\scriptsize\faIcon[regular]{address-card}}}}
\newcommand{\titletree}{%
  \mbox{\raisebox{-0.24em}{\includegraphics[height=1.25em]{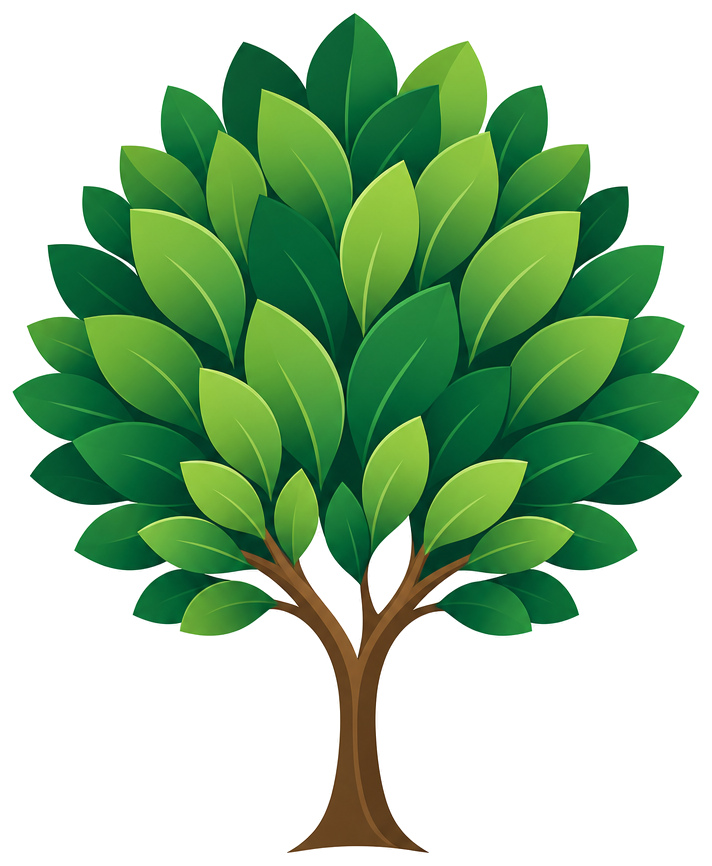}}}}
\tcbuselibrary{listings, breakable, skins}

\definecolor{shortbg}{HTML}{E3F2FD}
\definecolor{longbg}{HTML}{FFF3E0}
\newcommand{\shorttag}[1]{{\setlength{\fboxsep}{3pt}\colorbox{shortbg}{\strut #1}}}
\newcommand{\longtag}[1]{{\setlength{\fboxsep}{3pt}\colorbox{longbg}{\strut #1}}}
\definecolor{explicitbg}{HTML}{E8F5E9}
\definecolor{implicitbg}{HTML}{F3E5F5}
\newcommand{\explicittag}[1]{{\setlength{\fboxsep}{3pt}\colorbox{explicitbg}{\strut #1}}}
\newcommand{\implicittag}[1]{{\setlength{\fboxsep}{3pt}\colorbox{implicitbg}{\strut #1}}}
\definecolor{promptbg}{HTML}{F6F7F9}
\definecolor{promptframe}{HTML}{D4D8DE}
\newtcblisting{promptbox}{%
  enhanced, breakable, listing only,
  listing options={%
    basicstyle=\footnotesize\ttfamily,
    breaklines=true, columns=fullflexible, keepspaces=true,
    breakautoindent=false, prebreak={}, postbreak={},
    showspaces=false, showstringspaces=false, showtabs=false,
    upquote=true, extendedchars=true
  },
  colback=promptbg, colframe=promptframe,
  boxrule=0.4pt, arc=2pt, outer arc=2pt,
  left=8pt, right=6pt, top=5pt, bottom=5pt
}

\title{\texorpdfstring{\titletree\enspace}{}PHASE-Tree: Modeling Character-State Evolution in Long-Horizon Role-Playing Dialogue}

\author[1,*]{Bo Tang}
\author[1,*]{Jianan Yang}
\author[2]{Junyi Zhu}
\author[3]{Yiquan Wu}
\author[4]{Rui Zhao}
\author[5]{Zhengyu Yang}
\author[6]{Yang Zhang}
\author[1]{Feiyu Xiong}
\author[1,✉]{Zhiyu Li}
\author[1,✉]{Jiajun Shen}

\affiliation[1]{MemTensor (Shanghai) Technology}
\affiliation[2]{KU Leuven, Belgium}
\affiliation[3]{Zhejiang University}
\affiliation[4]{University of Chinese Academy of Sciences}
\affiliation[5]{Sinar Mas Paper (China) Investment Co., Ltd}
\affiliation[6]{The Hong Kong Polytechnic University}

\abstract{
Long-horizon role-playing demands that characters remain recognizable as they evolve with the narrative. Yet existing work falls short on two fronts: representations are typically static profiles that cannot be updated locally without destabilizing unchanged traits, and benchmarks mainly test persona preservation and memory recall rather than whether a model speaks from a character's currently evolved state. We address both. PHASE-Tree is a multi-timescale character-state tree with an immutable identity root and mutable persona, session, and moment layers, making each mutable field an addressable target for localized within- and cross-episode updates. It conditions generation through explicit textual provision or implicit parametric adaptation. To measure evolved-state generation, we introduce LongEvoRoleBench, which pairs four long-dialogue corpora for cross-episode evolution with four short-dialogue corpora as within-scene state-tracking checks, under a unified next-utterance protocol. On the long-dialogue core, textual PHASE-Tree ranks first in 11 of 12 dataset–metric cells against internal variants and all 12 cells against external textual baselines, improving character-level, semantic, and embedding scores by 19.7\%, 12.4\%, and 15.1\% respectively. In a blinded 200-response study, human ratings correlate with the GPT-4.1 judge (Pearson $r=0.65$); on descriptive $n=10$ PT and NR prompt subsets, the Overall difference is $+0.20$. The long-dialogue Sem advantage persists across LLM judges and generation backbones.
}
\checkdata[\iconcor\ Correspondence]{Zhiyu Li (\email{lizy@memtensor.cn}); Jiajun Shen (\email{sjjvic@gmail.com})}
\checkdata[\iconlegend\ Author Legend]{* Co-equal primary author, ✉ Corresponding authors}
\checkdata[\icongh\ Code]{\url{https://github.com/MemTensor/PHASE-Tree}}
\checkdata[\iconhf\ Dataset]{\url{https://huggingface.co/datasets/IAAR-Shanghai/LongEvoRoleBench}}
\checkdata[\iconhf\ Model]{\url{https://huggingface.co/IAAR-Shanghai/phase_tree_models}}

\begin{document}
\maketitle

\section{Introduction}
\label{sec:intro}
Long-horizon role-playing underpins interactive fiction, AI companions, and persistent game characters, where a model must remain recognizable while evolving with the narrative. Existing role-playing benchmarks and methods, however, mainly test whether a model preserves a fixed persona \citep{characterllm2023,rolellm2024,incharacter2024,coser2025} or recalls particular events \citep{locomo2024,personamem2025,persistentpersonas2026}, rather than whether it can generate from a character's currently evolved state. Realistic long-horizon role-playing requires more than preservation. Consider Chandler Bing in the television series \texttt{Friends}: early on he is sarcastic and commitment-phobic, but by later seasons he has grown into a husband who trusts his partner. A model that still treats commitment as a punchline in a marriage scene sounds superficially like Chandler while speaking from the wrong narrative state---the model has not forgotten his voice, but has forgotten that the character has changed. We call this \textbf{stale-state failure}.

This gap raises two questions. First, how should a character be represented so that the full richness of character state is expressible, yet individual attributes can be updated locally without destabilizing unchanged traits? Second, how can we evaluate whether a model generates from a character's currently evolved state across long narrative arcs, rather than regressing to a frozen persona?

We address both questions jointly. We propose \textbf{PHASE-Tree} (Psychology-grounded Hierarchical Attribute-Structured Evolving Tree), a multi-timescale character-state representation with an immutable identity root and mutable persona, session, and moment strata gated by a resistance--evidence--cooldown policy. To fill the evaluation gap, we introduce \textbf{LongEvoRoleBench}, a benchmark suite that standardizes eight existing role-playing corpora into a unified next-utterance protocol. Four long-dialogue corpora form the core test for cross-episode evolution, while four short-dialogue corpora provide within-scene state-tracking checks under the same evaluation format. Our PHASE-Tree representation can be consumed through two complementary conditioning paradigms: explicit textual provision, which serializes the tree into the prompt (our primary validated path), and implicit parametric adaptation (e.g., via a profile-to-LoRA hypernetwork~\citep{tan2025p2p}), a token-efficient alternative. 

Our contributions can be summarized as:
\begin{itemize}
    \item \textbf{PHASE-Tree character-state modeling.} A representation that decomposes character state into immutable identity facts and mutable persona, session, and moment attributes, with cross-episode evolution gated by resistance--evidence--cooldown policies.
    \item \textbf{LongEvoRoleBench.} A benchmark suite that standardizes eight role-playing corpora into a unified next-utterance protocol for evaluating both within-scene and cross-episode character-state evolution, with metrics tied to the current time-$t$ state rather than a frozen profile.
    \item \textbf{Systematic dual-paradigm validation.} We evaluate the same PHASE-Tree state under both explicit textual provision and implicit parametric adaptation, benchmarking against a comprehensive suite of ablation variants and external baselines. Our results show that textual provision achieves stronger alignment with evolved character states, while parametric adaptation is more token-efficient but reveals a compression bottleneck in current profile-to-LoRA architectures.
\end{itemize}

\section{Related Work}
\label{sec:related}

We summarize four related lines here; an extended discussion with a full citation list is given in Appendix~\ref{app:related-work-extended}.

\paragraph{Role-playing dialogue.}
Existing systems condition role-playing on static profiles, retrieved contexts, or per-character adapters \citep{characterllm2023,rolellm2024,oppu2024,tan2025p2p}, while evaluation-oriented work measures persona fidelity and established-role behavior \citep{incharacter2024,coser2025}. PHASE-Tree instead asks whether a role-playing model can speak from an appropriately evolved character state.
\paragraph{Long-horizon persona, memory, and drift.}
Recent dialogue systems and benchmarks treat persona and memory as dynamic and measure long-horizon fidelity and drift \citep{duelemon2022,ldagent2025,locomo2024,personamem2025,persistentpersonas2026,pdd2026}. PHASE-Tree narrows the claim to evolution-aware role-playing for fictional characters and updates editable fields through resistance, evidence, and cooldown gates.
\paragraph{Structured and psychology-grounded character modeling.}
Persona has been structured as trees, finite-state machines, or hierarchical identity frameworks, and shaped by psychological traits \citep{insideout2026,cfsm2026,identityhierarchical2025,profileaxes2026,chameleon2026}. PHASE-Tree separates an immutable identity root from three evolving time-scale strata and makes each editable field an update target.
\paragraph{Parameter-side personalization.}
Profile-to-LoRA hypernetworks, role-specific adapter generation, and activation-space steering have been studied for personalization and role-playing \citep{tan2025p2p,hyperlora2024,neeko2024,hycora2026,persona2026}. We treat implicit parametric adaptation as a complementary, token-efficient variant; the main validated path is explicit textual provision.

Closest in spirit are the user-memory and persona-fidelity benchmarks cited above together with \citet{horizonbench2026}; LongEvoRoleBench differs by evaluating fictional role characters whose narrative state changes are part of the task, with metrics and baselines for both textual-provision and parametric-adaptation paradigms.

\section{PHASE-Tree: Character-State Evolution Modeling}
\label{sec:method}

\begin{figure}[htbp]
    \centering
    \includegraphics[width=\columnwidth]{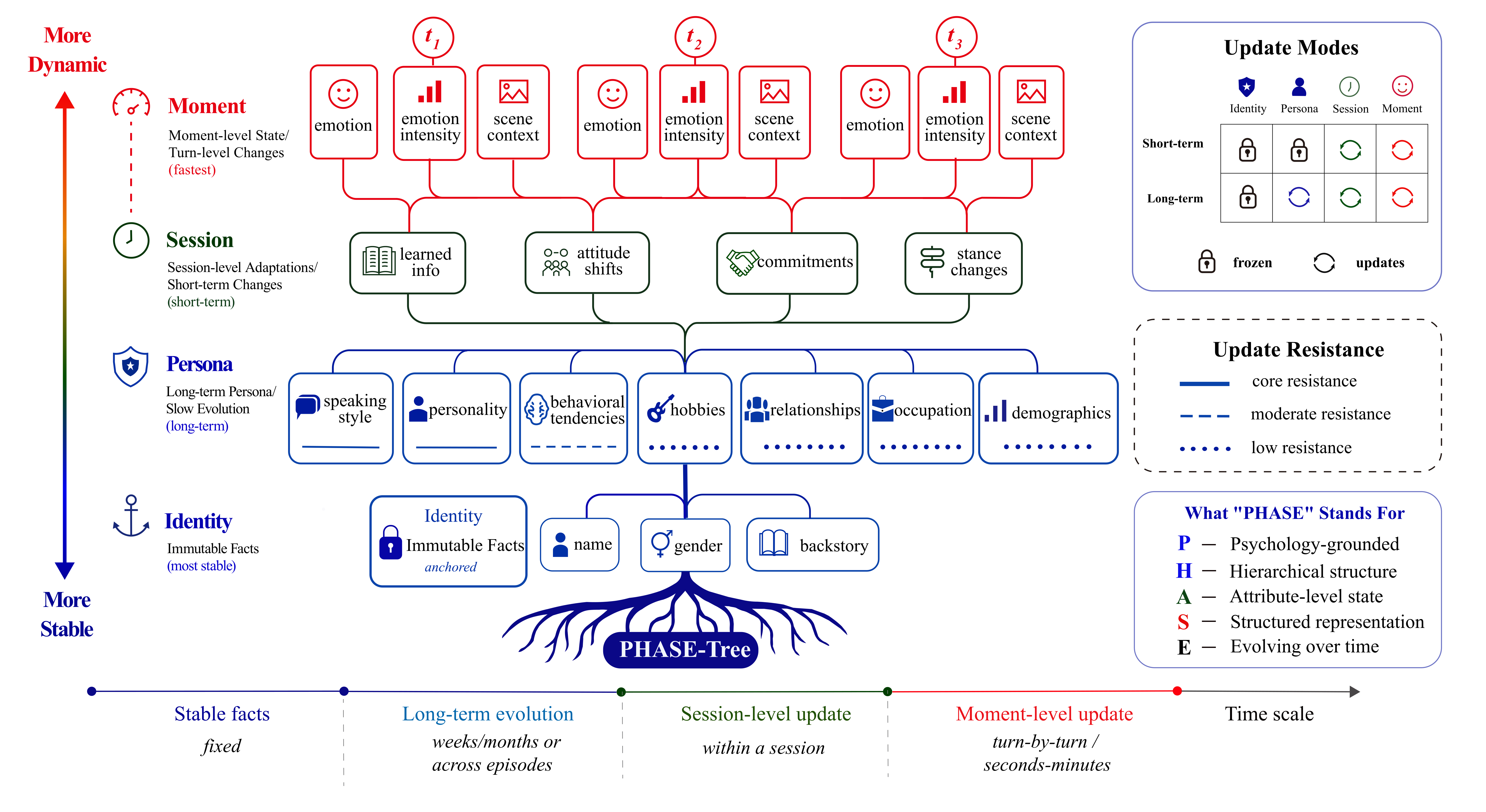}
    \caption{PHASE-Tree character-state hierarchy. An immutable identity root branches into persona, session, and moment strata; persona fields use resistance levels (solid = core, dashed = moderate, dotted = low).}
    \label{fig:phase-tree}
\end{figure}

A static profile cannot fully capture a believable character: some attributes (name, gender, and backstory) never change, others (speaking style and personality) change slowly under sustained evidence, and still others (mood and stance) may shift within a scene. PHASE-Tree encodes these time scales in a four-stratum tree with an immutable identity root and three mutable strata. Figure~\ref{fig:phase-tree} gives an overview.

\subsection{Character State and Update Dynamics}
\label{sec:formulation}

Suppose a target character appears in $T$ dialogue scenes. For scene $t$, a contiguous block of $n_t$ turns, the previous turns $c_t = (u_{t,1}, u_{t,2}, \ldots, u_{t,n_t-1})$ constitute the dialogue context, and the next turn $y_t = u_{t,n_t}$ is the target character's ground-truth response. For long-dialogue corpora, scenes are grouped into episodes and seasons, yielding a common season--episode--scene hierarchy. Scene-level evidence drives local state tracking, while episode boundaries govern long-term persona evolution.

We model the character state at time $t$ as a four-part structured tree:
\begin{equation}
    \mathcal{S}_t = \big(\, \mathcal{I},\; \mathcal{S}^{\text{persona}}_t,\; \mathcal{S}^{\text{session}}_t,\; \mathcal{S}^{\text{moment}}_t \,\big)
\end{equation}
where $\mathcal{I}$ stores immutable identity facts (name, gender, and backstory), and the three mutable strata are:\footnote{We use ``character state'' to denote this role-playing mental and social state, rather than game-state variables such as inventory or health.}

\begin{itemize}
    \item \textbf{Persona} $\mathcal{S}_t^{\mathrm{persona}}$ (\textit{high resistance}): 
    long-term dispositions and relatively stable profile attributes (personality, speaking style, 
    behavioral tendencies, hobbies, relationships, occupation, demographics).
    
    \item \textbf{Session} $\mathcal{S}_t^{\mathrm{session}}$ (\textit{moderate resistance}): 
    within-scene characteristic adaptations (newly learned information, attitude shifts, 
    commitments, and stance changes) accumulated during the current scene.
    
    \item \textbf{Moment} $\mathcal{S}_t^{\mathrm{moment}}$ (\textit{low resistance}): 
    transient state affect, specifically the dominant emotion, its intensity, and the 
    triggering scene context, refreshed at each scene boundary~\citep{spielberger1983manual,rosenberg1998levels}.
\end{itemize}

The persona--session distinction is motivated by McAdams' separation of broad dispositional traits from contextualized characteristic adaptations \citep{mcadams1995what}, while the moment stratum follows the state--trait distinction in affect psychology \citep{spielberger1983manual,rosenberg1998levels}. The full schema with psychological grounding is given in Appendix~\ref{app:schema}. Every editable field is independently addressable: an update targets one field without rewriting siblings. After each scene, an extraction function $\mathcal{E}$ produces evidence $e_t = \mathcal{E}(c_t, y_t, \mathcal{S}_t)$, and $\mathcal{S}_{t+1} = \mathcal{U}(\mathcal{S}_t, e_t)$. The update function modifies only the mutable strata and leaves $\mathcal{I}$ unchanged. Thus $\mathcal{S}_t$ conditions $y_t$, whose evidence first enters $\mathcal{S}_{t+1}$.

We have defined the state formalism and its update dynamics. We now turn to how this tree is initially instantiated from raw character profiles.

\subsection{Tree Construction}
\label{sec:tree-construction}

We assume an initial raw character profile is available and map it once into PHASE-Tree with a fixed zero-shot GPT-4.1 extractor. The extractor parses heterogeneous source fields into the standardized identity and persona schema, while initializing session and moment to empty/default values. The same prompt template is applied across all eight corpora without corpus-specific manual authoring or rule engineering. Field values remain free-text rather than categorical, preserving character-specific detail within a uniform structure. This one-time preprocessing produces a baseline tree that can be cached and reused across inference calls.

This baseline state is the starting point: the session and moment strata must be updated to track developments within each scene, and the persona stratum evolves across episodes as the character undergoes lasting change.

\subsection{Intra-Scene State Tracking}
\label{sec:tracking}

Even within a single scene, a character's response should reflect what they have learned and felt in the preceding dialogue: discovering a betrayal should reshape the character's subsequent stance. For each target turn $y_t$, an LLM analyzes only the observed prefix $c_t$ together with the scene-start identity and persona. It extracts (i) a third-person session entry covering newly learned information, attitude shifts, commitments, and stance changes, and (ii) a moment snapshot capturing the dominant emotion, its intensity, and the current scene context. These session and moment fields condition $y_t$; evidence from $y_t$ first enters the state used by a later target. When the scene closes, the local records are archived as evidence for subsequent persona evolution. Multi-character scenes use independent extraction passes for each main character.

\subsection{Cross-Episode Persona Evolution}
\label{sec:evolution}
In long-horizon narratives, a character's long-term traits can genuinely change. Chandler Bing's commitment-avoidance, for example, gives way to marital 
responsibility over several seasons. Yet not all persona fields evolve at the same rate: a relationship may end from a single decisive scene, whereas a core personality shift requires sustained evidence across many episodes. We capture this differential plasticity through a three-stage pipeline that updates persona fields independently, with per-field resistance calibrated to narrative pacing.

\paragraph{Stage 1: Evidence Accumulation.}
Independently of the per-scene session/moment extraction in \S\ref{sec:tracking}, a separate LLM pass scans each scene from the character's perspective, identifies salient session events (e.g., a reconciliation with an estranged partner or a career-changing commitment), and labels each with a significance level (medium or high). The labeled entries are appended to the character's session archive and serve as the evidence base for subsequent evolution decisions.

\paragraph{Stage 2: Resistance-Gated Judgment.}
After each episode, an LLM proposes per-field candidate updates given the active evidence archive, and a deterministic validator accepts or rejects each proposal under threshold checks. Every evolvable field $f$ carries a resistance level $r(f) \in \{\text{core}, \text{moderate}, \text{low}\}$. Three complementary checks prevent premature or unstable updates: breadth ($n_{\text{ep}}$) ensures the change is visible across multiple episodes rather than a single anomalous scene, intensity ($n_{\text{high}}$) requires at least some high-significance events to support it, and cooldown ($\Delta_{\text{ep}}$) requires enough elapsed episodes since the last update to the same field, preventing rapid flip-flopping. Formally, a field is updated only when all three conditions hold jointly:
\begin{equation}
\mathrm{update}(f) \iff n_{\mathrm{ep}}(f) \geq \tau^{\mathrm{ep}}_{r(f)} \wedge\; n_{\mathrm{high}}(f) \geq \tau^{\mathrm{high}}_{r(f)} \wedge\; \Delta_{\mathrm{ep}}(f) \geq \tau^{\mathrm{cd}}_{r(f)}.
\end{equation}
where $n_{\mathrm{ep}}(f)$ is the number of distinct episodes contributing evidence to $f$, $n_{\mathrm{high}}(f)$ is the number of high-significance entries, and $\Delta_{\mathrm{ep}}(f)$ is the number of episodes elapsed since the last update to $f$. Higher resistance imposes stricter thresholds along all three axes; for instance, $\tau^{\mathrm{ep}}_{\text{core}} > \tau^{\mathrm{ep}}_{\text{moderate}} > \tau^{\mathrm{ep}}_{\text{low}}$, and the same ordering holds for $\tau^{\mathrm{high}}$ and $\tau^{\mathrm{cd}}$. Concretely, \texttt{personality} and \texttt{speaking\_style} are core fields (requiring evidence from $\geq$16 episodes with $\geq$6 high-significance entries), \texttt{behavioral\_tendencies} is moderate ($\geq$3 episodes), and \texttt{relationships}, \texttt{occupation}, \texttt{hobbies}, and \texttt{demographics} are low (1 high-significance entry or 2 medium-significance entries suffice; for the low tier, this disjunctive evidence rule overrides the generic $\tau^{\mathrm{ep}}$ episode count). Thus core traits demand evidence spanning roughly a full season, whereas a relationship status can update from a single decisive event. The thresholds are manually set once from narrative-pacing priors and held fixed across all four long-dialogue corpora, without corpus-specific tuning. Learning them automatically is left to future work. Full threshold values appear in Appendix~\ref{app:evolution-params}.

\paragraph{Stage 3: Incremental Field Update.}
When the gating conditions are met, the system applies a single-field update using one of two merge operations: an \textsc{incremental} merge that adds or refines content while preserving the previous value (covering both list-append and in-place refinement of free-text fields), or a \textsc{replacement} merge that substitutes the previous value entirely and is reserved for explicit contradictions. 
A small set of post-processing patches then handles edge cases that arise in multi-character corpora: stale relationship entries are demoted when they lack recent evidence, reciprocity gaps between interacting characters are repaired, and continuity is forward-filled to avoid regression across sequential episodes. The full pipeline, together with a human audit of the accepted field updates, is described in Appendix~\ref{app:pipeline}; the extraction and update prompt specifications are provided in Appendix~\ref{app:prompts}.

\subsection{Generation Conditioning Paradigms}
\label{sec:paradigm}
\begin{figure}[htbp]
    \centering
    \begin{minipage}[c]{0.65\textwidth}
        \centering
        \includegraphics[width=\linewidth]{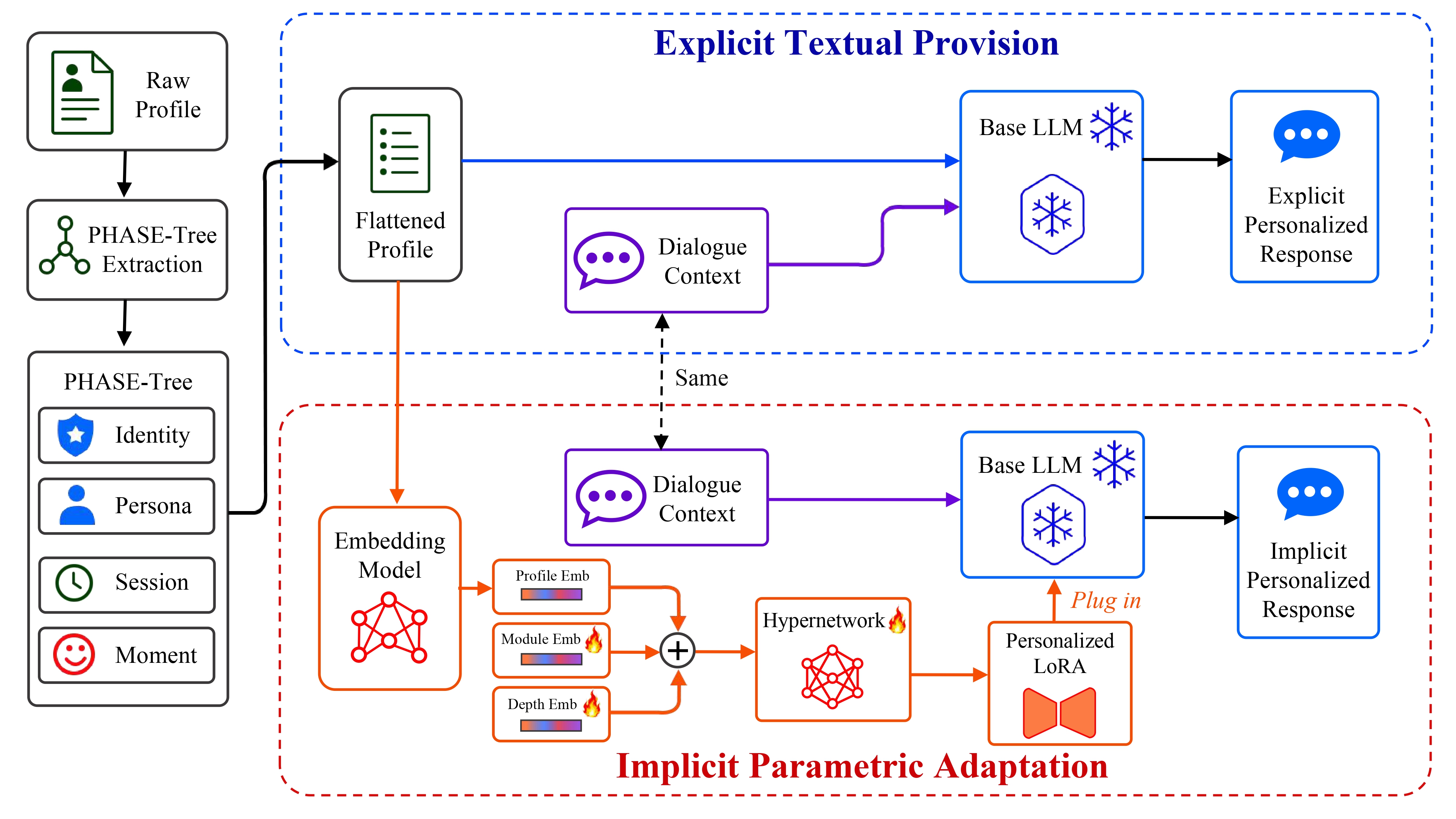}
    \end{minipage}\hfill
    \begin{minipage}[c]{0.34\textwidth}
        \caption{Two conditioning paradigms for the same flattened PHASE-Tree. \textbf{Explicit Textual Provision} (blue, top): profile in the prompt. \textbf{Implicit Parametric Adaptation} (red, bottom): profile in adapter weights; dialogue-only prompt. \textcolor{red}{\textbf{Fire}} indicate trainable components; 
        \textcolor{blue}{\textbf{Snowflake}} indicate frozen parameters.}
        \label{fig:training-pipeline}
    \end{minipage}
\end{figure}

We implement two complementary conditioning paradigms that consume the character state $\mathcal{S}_t$ at inference time, illustrated side-by-side in 
Figure~\ref{fig:training-pipeline}.

\paragraph{Explicit Textual Provision.}
The tree is serialized into structured natural-language paragraphs (identity facts, persona traits, session adaptations, and momentary affect) and concatenated with the dialogue context:
\begin{equation}
    \hat{y}_t = \arg\max_{y}\; p_{\theta}\!\big(y \mid \mathrm{flatten}(\mathcal{S}_t),\; c_t\big),
\end{equation}
where $\mathrm{flatten}(\cdot)$ converts the tree into the above natural-language 
groups.

\paragraph{Implicit Parametric Adaptation.}
Following profile-to-LoRA hypernetworks \citep{tan2025p2p}, a hypernetwork $H_\phi$ 
maps the embedded character state to a LoRA adapter 
$\Delta\theta_t = H_\phi(\mathrm{emb}(\mathcal{S}_t))$ that is merged into the backbone. The prompt then carries only dialogue context:
\begin{equation}
    \hat{y}_t = \arg\max_{y}\; p_{\theta + \Delta\theta_t}\!\big(y \mid c_t\big),
\end{equation}
where $\mathrm{emb}(\cdot)$ encodes the flattened tree via an embedding model. 
The adapter $\Delta\theta_t$ is state-dependent and changes whenever 
$\mathcal{S}_t$ evolves; holding $\mathcal{S}_t$ constant recovers a static 
profile-to-LoRA mapping.

\section{LongEvoRoleBench: A Benchmark for Long-Horizon Role-Playing}
\label{sec:benchmark}

Existing role-playing benchmarks typically evaluate persona preservation or memory recall against a fixed profile, failing to test if a model can accurately reflect a character's evolving state. To fill this gap, we introduce LongEvoRoleBench, which evaluates a system's ability to ground a character's next utterance in their current narrative state. The benchmark comprises eight datasets evaluated under unified metrics: four long-dialogue corpora constitute the core test of cross-episode character evolution, while four short-dialogue corpora serve as a control setting that isolates within-scene consistency and local state tracking without cross-episode evolution.

\subsection{Dataset Construction}
\label{sec:suite-construction}

LongEvoRoleBench unifies eight role-playing corpora under a common next-utterance protocol with standardized profile schemas, context boundaries, and evaluation splits. Each instance provides scene context $c_t$ and targets the character's next utterance $y_t$. Short-dialogue sources are RAIDEN~\citep{raiden2025}, CharacterEval (RPCA benchmarks)~\citep{charactereval2024}, SimsConv~\citep{simsconv2025} and ChatHaruhi~\citep{chatharuhi2023}. Long-dialogue sources are Friends (ConvoKit; \citealp{convokit2020}), The Office and Star Trek (public episode transcripts), and Harry Potter (HPD; \citealp{hpd2023}), each tracking six main characters whose relationships and affect evolve across seasons or books. Corpus statistics are in Appendix~\ref{app:datasets}.

\subsection{Evaluation Protocol}
\label{sec:protocol}

We evaluate under complementary random and OOD holdouts. Short-dialogue splits are character-disjoint: random tests sample held-out characters from the main profile distribution, while OOD tests select profile outliers through embedding clustering, probing generalization to less familiar character types. Long-dialogue random tests withhold utterance instances from early seasons, whereas OOD tests chronologically hold out later seasons, where relationships, beliefs, and affect may have evolved substantially. Together, these splits distinguish interpolation within familiar narrative regimes from extrapolation across character types and narrative time. Character states roll forward in narrative order, combining the initial profile with preceding scene and episode evidence at each prediction point. All parametric baselines are trained and all retrieval indices are built on the train split, and scoring uses the matching time-$t$ state rather than a frozen profile.

\subsection{Evaluation Metrics}
\label{sec:metrics}
Generation quality is measured along three complementary axes. Character Score (Char) and Semantic Score (Sem) are independent 1--5 LLM-as-Judge ratings (GPT-4.1, greedy decoding) for profile consistency and contextual coherence. The judge evaluates responses using the supplied profile text (Char) or dialogue context (Sem). Embedding Score (Emb) is cosine similarity to the ground-truth response under OpenAI's \texttt{text-embedding-3-small}. The three scores target distinct failure modes: profile drift, contextual mismatch, and reference-grounded semantic distance. A strong method must satisfy all three rather than win on one alone. We omit n-gram metrics (BLEU, ROUGE) because long-horizon role-playing admits many surface-divergent yet equally valid continuations. Appendices~\ref{app:judge} and~\ref{app:judge-validity} provide the scoring rubrics and judge analyses.

\section{Experiments}
\label{sec:results}

\subsection{Experimental Setup}
\label{sec:impl}

The main comparison uses Qwen2.5-7B-Instruct as the shared backbone, so methods differ only in how the character condition is supplied. Explicit textual provision keeps the backbone frozen and places the condition in the prompt; implicit parametric adaptation encodes it through network adapters or inference-time activation vectors.

\paragraph{Training and Decoding.} All runs use a fixed decoding configuration (temperature 0.3, max 256 tokens, seed 42). PHASE-Trees are constructed by the extraction pipeline in \S\ref{sec:tree-construction} using GPT-4.1. Our implicit parametric-adaptation route warm-starts from a P2P checkpoint \citep{tan2025p2p} and is further finetuned end-to-end. Full implementation details are in Appendix~\ref{app:exp-impl}.

\paragraph{Baselines.} We first run a progressive internal ablation that adds one representation component at a time to isolate its contribution: \textbf{Base} (no profile), \textbf{RP} (raw profile), \textbf{NR} (LLM-rewritten profile), \textbf{ST} (structured tree, frozen), \textbf{DT} (tree with cross-episode persona evolution, no session/moment), and \textbf{PT} (full pipeline with intra-episode tracking). The external baselines span both conditioning paradigms. Textual-provision baselines are \textbf{RAG} (retrieved historical utterances; \citealp{lewis2020rag}), \textbf{PAG} (profile-augmented prompt; \citealp{characterllm2023,rolellm2024,chatharuhi2023}), and \textbf{CFG} (decoding-time amplification; \citealp{sanchez2024cfg}). Parametric-adaptation baselines are \textbf{MT-LoRA} (a single shared LoRA adapter trained on the union of all character dialogues; \citealp{hu2022lora}), \textbf{Steering} (activation vectors; \citealp{turner2024activation}), \textbf{OPPU} (per-character adapter, \citealp{oppu2024}), and \textbf{P2P} (profile-to-LoRA hypernetwork, \citealp{tan2025p2p}). Per-method details and controls for fair comparison are presented in Appendices~\ref{app:ablation-chain} and \ref{app:external-baselines}.

\begin{table}[t!]
  \centering
  \small
  \caption{Internal ablation under explicit textual provision on eight corpora (mean over random and OOD splits). \textbf{Bold} = best, \underline{underline} = second best; DT applies to long-dialogue sets only (--). Pooled question-level tests and effect sizes for PT vs.\ NR and PT vs.\ ST are reported in Appendix~\ref{app:stats}.}
  \label{tab:explicit-full}
  \begin{tabular}{ll ccccc c}
  \toprule
  \textbf{Dataset} & \textbf{Metric} & \textbf{Base} & \textbf{RP} & \textbf{NR} & \textbf{ST} & \textbf{DT} & \textbf{PT (Ours)} \\
  \midrule
  \multirow{3}{*}{\shorttag{RAIDEN}}
   & Char $\uparrow$ & 2.163 & 2.691 & 2.736 & \underline{2.769} & -- & \textbf{2.779} \\
   & Sem $\uparrow$  & 3.632 & 3.566 & \underline{3.812} & 3.752 & -- & \textbf{3.845} \\
   & Emb $\uparrow$  & 0.444 & 0.442 & 0.463 & \underline{0.464} & -- & \textbf{0.468} \\
  \midrule
  \multirow{3}{*}{\shorttag{CharacterEval}}
   & Char $\uparrow$ & 2.188 & 2.474 & 2.753 & \underline{2.785} & -- & \textbf{2.790} \\
   & Sem $\uparrow$  & 3.382 & 3.202 & \underline{3.560} & 3.455 & -- & \textbf{3.588} \\
   & Emb $\uparrow$  & \underline{0.325} & 0.314 & 0.321 & 0.323 & -- & \textbf{0.326} \\
  \midrule
  \multirow{3}{*}{\shorttag{SimsConv}}
   & Char $\uparrow$ & 2.339 & \underline{3.206} & 2.960 & \textbf{3.294} & -- & 3.093 \\
   & Sem $\uparrow$  & 3.749 & 3.819 & \underline{3.881} & 3.825 & -- & \textbf{3.927} \\
   & Emb $\uparrow$  & 0.439 & 0.446 & \underline{0.454} & 0.443 & -- & \textbf{0.466} \\
  \midrule
  \multirow{3}{*}{\shorttag{ChatHaruhi}}
   & Char $\uparrow$ & 1.880 & 3.134 & \textbf{3.606} & 3.424 & -- & \underline{3.451} \\
   & Sem $\uparrow$  & 3.391 & 3.307 & \underline{3.764} & 3.693 & -- & \textbf{3.810} \\
   & Emb $\uparrow$  & 0.367 & 0.398 & 0.419 & \underline{0.420} & -- & \textbf{0.425} \\
  \midrule
  \multirow{3}{*}{\longtag{Friends}}
   & Char $\uparrow$ & 2.304 & 2.371 & \underline{2.860} & 2.794 & 2.757 & \textbf{2.907} \\
   & Sem $\uparrow$  & 3.303 & 2.679 & \underline{3.443} & 3.293 & 3.309 & \textbf{3.650} \\
   & Emb $\uparrow$  & 0.262 & 0.227 & \underline{0.265} & 0.260 & 0.261 & \textbf{0.298} \\
  \midrule
  \multirow{3}{*}{\longtag{The Office}}
   & Char $\uparrow$ & 2.102 & 2.622 & \textbf{3.086} & \underline{3.061} & 3.051 & 3.007 \\
   & Sem $\uparrow$  & 3.368 & 2.676 & \underline{3.465} & 3.317 & 3.377 & \textbf{3.756} \\
   & Emb $\uparrow$  & \underline{0.254} & 0.225 & 0.250 & 0.245 & 0.247 & \textbf{0.293} \\
  \midrule
  \multirow{3}{*}{\longtag{Harry Potter}}
   & Char $\uparrow$ & 2.342 & 2.477 & \underline{2.930} & 2.782 & 2.808 & \textbf{2.961} \\
   & Sem $\uparrow$  & 3.257 & 2.904 & \underline{3.462} & 3.330 & 3.355 & \textbf{3.636} \\
   & Emb $\uparrow$  & 0.273 & 0.256 & \underline{0.292} & 0.287 & 0.285 & \textbf{0.322} \\
  \midrule
  \multirow{3}{*}{\longtag{Star Trek}}
   & Char $\uparrow$ & 2.557 & 2.345 & \underline{3.080} & 2.980 & 2.957 & \textbf{3.139} \\
   & Sem $\uparrow$  & 3.363 & 2.837 & \underline{3.526} & 3.390 & 3.410 & \textbf{3.746} \\
   & Emb $\uparrow$  & 0.283 & 0.266 & \underline{0.294} & 0.292 & 0.291 & \textbf{0.343} \\
  \midrule
  \multicolumn{8}{c}{\textit{Average Performance}} \\
  \midrule
  \multirow{3}{*}{\shorttag{Short-Dialogue}}
   & Char $\uparrow$ & 2.143 & 2.876 & 3.014 & \textbf{3.068} & -- & \underline{3.028} \\
   & Sem $\uparrow$  & 3.539 & 3.474 & \underline{3.754} & 3.681 & -- & \textbf{3.792} \\
   & Emb $\uparrow$  & 0.394 & 0.400 & \underline{0.414} & 0.412 & -- & \textbf{0.421} \\
  \midrule
  \multirow{3}{*}{\longtag{Long-Dialogue}}
   & Char $\uparrow$ & 2.326 & 2.454 & \underline{2.989} & 2.904 & 2.894 & \textbf{3.004} \\
   & Sem $\uparrow$  & 3.323 & 2.774 & \underline{3.474} & 3.332 & 3.363 & \textbf{3.697} \\
   & Emb $\uparrow$  & 0.268 & 0.244 & \underline{0.275} & 0.271 & 0.271 & \textbf{0.314} \\
  \bottomrule
  \end{tabular}
  \end{table}

  \begin{table}[t!]
    \centering
    \small
    \setlength{\tabcolsep}{3pt}
    \caption{External baseline comparison on eight corpora (mean over random and OOD splits). \textbf{Bold} = best, \underline{underline} = second best. Ours denotes PHASE-Tree in the corresponding paradigm block; -- marks unavailable runs. Pooled question-level tests and effect sizes for key comparisons are reported in Appendix~\ref{app:stats}.}
    \label{tab:external-full}
    \begin{tabular}{ll ccccc !{\vrule width 0.6pt} ccccc}
    \toprule
    & & \multicolumn{5}{c!{\vrule width 0.6pt}}{\textit{Explicit Textual Provision}} & \multicolumn{5}{c}{\textit{Implicit Parametric Adaptation}} \\
    \cmidrule(lr){3-7}\cmidrule(lr){8-12}
    \textbf{Dataset} & \textbf{Metric} & \textbf{Base} & \textbf{RAG} & \textbf{PAG} & \textbf{CFG} & \textbf{Ours} & \textbf{MT-LoRA} & \textbf{Steering} & \textbf{OPPU} & \textbf{P2P} & \textbf{Ours} \\
    \midrule
    \multirow{3}{*}{\shorttag{RAIDEN}}
     & Char $\uparrow$ & 2.163 & 2.511 & \underline{2.842} & \textbf{2.854} & 2.779 & \underline{2.506} & 2.202 & -- & 2.492 & \textbf{2.510} \\
     & Sem $\uparrow$  & 3.632 & \underline{3.738} & 3.582 & 3.347 & \textbf{3.845} & \textbf{3.922} & 3.639 & -- & 3.894 & \underline{3.915} \\
     & Emb $\uparrow$  & 0.444 & \textbf{0.468} & 0.452 & 0.423 & \underline{0.468} & \textbf{0.505} & 0.445 & -- & 0.487 & \underline{0.505} \\
    \midrule
    \multirow{3}{*}{\shorttag{CharacterEval}}
     & Char $\uparrow$ & 2.188 & 2.453 & \underline{2.597} & 2.573 & \textbf{2.790} & 2.312 & 2.170 & -- & \textbf{2.341} & \underline{2.334} \\
     & Sem $\uparrow$  & 3.382 & \underline{3.475} & 3.362 & 2.981 & \textbf{3.588} & \textbf{3.550} & 3.391 & -- & 3.519 & \underline{3.548} \\
     & Emb $\uparrow$  & 0.325 & \textbf{0.360} & \underline{0.341} & 0.298 & 0.326 & \textbf{0.346} & 0.323 & -- & 0.337 & \underline{0.345} \\
    \midrule
    \multirow{3}{*}{\shorttag{SimsConv}}
     & Char $\uparrow$ & 2.339 & 2.788 & \underline{3.223} & \textbf{3.565} & 3.093 & \underline{2.489} & 2.342 & -- & \textbf{2.503} & 2.455 \\
     & Sem $\uparrow$  & 3.749 & \underline{3.876} & 3.864 & 3.590 & \textbf{3.927} & \underline{3.944} & 3.808 & -- & 3.918 & \textbf{3.977} \\
     & Emb $\uparrow$  & 0.439 & \textbf{0.489} & \underline{0.477} & 0.415 & 0.466 & \underline{0.515} & 0.444 & -- & 0.496 & \textbf{0.523} \\
    \midrule
    \multirow{3}{*}{\shorttag{ChatHaruhi}}
     & Char $\uparrow$ & 1.880 & 2.354 & 3.294 & \underline{3.308} & \textbf{3.451} & 1.893 & 1.888 & -- & \underline{1.950} & \textbf{1.976} \\
     & Sem $\uparrow$  & 3.391 & \underline{3.546} & 3.544 & 3.063 & \textbf{3.810} & \underline{3.527} & 3.379 & -- & 3.492 & \textbf{3.551} \\
     & Emb $\uparrow$  & 0.367 & 0.418 & \textbf{0.436} & 0.387 & \underline{0.425} & \textbf{0.414} & 0.366 & -- & 0.386 & \underline{0.412} \\
    \midrule
    \multirow{3}{*}{\longtag{Friends}}
     & Char $\uparrow$ & 2.304 & 2.361 & \underline{2.417} & 2.386 & \textbf{2.907} & 2.170 & \textbf{2.589} & 2.308 & \underline{2.379} & 2.205 \\
     & Sem $\uparrow$  & \underline{3.303} & 3.293 & 2.905 & 2.378 & \textbf{3.650} & \textbf{3.417} & 2.649 & 3.306 & 3.371 & \underline{3.416} \\
     & Emb $\uparrow$  & 0.262 & \underline{0.273} & 0.250 & 0.212 & \textbf{0.298} & \textbf{0.279} & 0.241 & 0.262 & 0.269 & \underline{0.278} \\
    \midrule
    \multirow{3}{*}{\longtag{The Office}}
     & Char $\uparrow$ & 2.102 & 2.175 & \underline{2.647} & 2.617 & \textbf{3.007} & 2.050 & \underline{2.235} & \textbf{2.255} & 2.117 & 2.112 \\
     & Sem $\uparrow$  & \underline{3.368} & 3.306 & 2.740 & 2.350 & \textbf{3.756} & \underline{3.476} & 1.759 & 3.151 & 3.449 & \textbf{3.498} \\
     & Emb $\uparrow$  & \underline{0.254} & 0.252 & 0.228 & 0.204 & \textbf{0.293} & \textbf{0.268} & 0.227 & 0.261 & 0.260 & \underline{0.267} \\
    \midrule
    \multirow{3}{*}{\longtag{Harry Potter}}
     & Char $\uparrow$ & 2.342 & 2.439 & \underline{2.476} & 2.403 & \textbf{2.961} & 2.325 & \underline{2.443} & \textbf{2.497} & 2.419 & 2.381 \\
     & Sem $\uparrow$  & \underline{3.257} & 3.234 & 2.963 & 2.507 & \textbf{3.636} & \underline{3.380} & 2.549 & 3.052 & 3.350 & \textbf{3.396} \\
     & Emb $\uparrow$  & 0.273 & \underline{0.280} & 0.268 & 0.237 & \textbf{0.322} & 0.288 & 0.268 & \textbf{0.312} & 0.280 & \underline{0.290} \\
    \midrule
    \multirow{3}{*}{\longtag{Star Trek}}
     & Char $\uparrow$ & 2.557 & \underline{2.644} & 2.498 & 2.149 & \textbf{3.139} & \underline{2.533} & 2.257 & 2.444 & \textbf{2.670} & 2.528 \\
     & Sem $\uparrow$  & \underline{3.363} & 3.322 & 2.946 & 2.482 & \textbf{3.746} & \underline{3.439} & 2.444 & 3.054 & \textbf{3.469} & 3.426 \\
     & Emb $\uparrow$  & 0.283 & \underline{0.285} & 0.274 & 0.249 & \textbf{0.343} & 0.296 & 0.259 & 0.296 & \underline{0.296} & \textbf{0.297} \\
    \midrule
    \multicolumn{12}{c}{\textit{Average Performance}} \\
    \cmidrule(lr){1-12}
    \multirow{3}{*}{\shorttag{Short-Dialogue}}
     & Char $\uparrow$ & 2.143 & 2.527 & 2.989 & \textbf{3.075} & \underline{3.028} & 2.300 & 2.150 & -- & \textbf{2.321} & \underline{2.319} \\
     & Sem $\uparrow$  & 3.539 & \underline{3.659} & 3.588 & 3.245 & \textbf{3.792} & \underline{3.736} & 3.554 & -- & 3.706 & \textbf{3.748} \\
     & Emb $\uparrow$  & 0.394 & \textbf{0.434} & \underline{0.427} & 0.381 & 0.421 & \underline{0.445} & 0.394 & -- & 0.427 & \textbf{0.446} \\
    \midrule
    \multirow{3}{*}{\longtag{Long-Dialogue}}
     & Char $\uparrow$ & 2.326 & 2.405 & \underline{2.510} & 2.389 & \textbf{3.004} & 2.269 & \underline{2.381} & 2.376 & \textbf{2.396} & 2.306 \\
     & Sem $\uparrow$  & \underline{3.323} & 3.289 & 2.889 & 2.429 & \textbf{3.697} & \underline{3.428} & 2.350 & 3.141 & 3.410 & \textbf{3.434} \\
     & Emb $\uparrow$  & 0.268 & \underline{0.273} & 0.255 & 0.225 & \textbf{0.314} & 0.283 & 0.249 & \underline{0.283} & 0.276 & \textbf{0.283} \\
    \bottomrule
    \end{tabular}
    \end{table}

\subsection{Results}
\label{sec:results-main}

\paragraph{Explicit Textual Provision.}
Across the full per-dataset breakdown (Table~\ref{tab:explicit-full}), PT ranks first on all eight datasets for Sem and Emb and on five of eight for Char, yielding the best score in 21 of 24 dataset--metric cells. On the four long-dialogue corpora, PT leads 11 of 12 cells; NR exceeds it only on The Office Char by $0.079$. The three Char exceptions overall (SimsConv, ChatHaruhi, and The Office) arise when short or highly stylized source profiles make a closer surface paraphrase (NR or ST) match the LLM judge's lexical expectations more readily than a restructured tree; Appendix~\ref{app:judge-validity} isolates this reference sensitivity. 

Progressive ablations support three conclusions. PT vs.\ NR suggests that separating immutable identity information from mutable state provides cleaner conditioning. ST and DT remain below NR on all long-dialogue macro-averages, so structure or cross-episode evolution alone is insufficient. DT to PT is the only structured transition that improves all three metrics (+0.110 Char, +0.334 Sem, and +0.043 Emb), showing that session and moment layers supply transient cues missed by cross-episode evolution alone.

The cross-backbone analysis further evaluates Qwen3-0.6B, Gemma-4-E4B, Qwen2.5-7B, and Qwen3-32B. PT achieves the best long-dialogue Sem on all four backbones, with the full results reported in Appendix~\ref{app:backbone}.

\paragraph{Comparison with External Methods.}
Under textual provision, Ours ranks first in all 12 long-dialogue dataset--metric cells, first on 18 of 24 cells overall, and in the top two on 20 (Table~\ref{tab:external-full}). The average improvement over the strongest textual-provision baseline for each long-dialogue metric is +0.49~Char (3.00 vs.\ PAG's 2.51, +19.7\%), +0.41~Sem (3.70 vs.\ RAG's 3.29, +12.4\%), and +0.04~Emb (0.31 vs.\ RAG's 0.27, +15.1\%). 
Because RAG, PAG, and CFG share the same backbone and evaluation pipeline, these comparisons isolate the effect of profile representation from major implementation differences.

\paragraph{Implicit Parametric Adaptation.}
The implicit route feeds the same flattened PHASE-Tree to a profile-to-LoRA hypernetwork, absorbing the profile into adapter weights so the generation prompt contains only dialogue context. Within the parametric-adaptation block of Table~\ref{tab:external-full}, Ours ranks first on 8 of 24 dataset--metric cells and in the top two on 18, leading on Sem in both short-dialogue and long-dialogue averages and tying for first on long-dialogue Emb. However, within the internal ablation, RP, NR, ST, and PT are very close on both Sem and Emb (most rows within $\pm$0.01; full results in Appendix~\ref{app:implicit-results}), and Char remains lower than under textual provision. This pattern indicates that the bottleneck lies in the profile-to-LoRA mapping, which compresses away the fine-grained state detail that distinguishes tree variants, rather than in the input representation.

\begin{figure}[t!]
  \centering
  \includegraphics[width=\columnwidth]{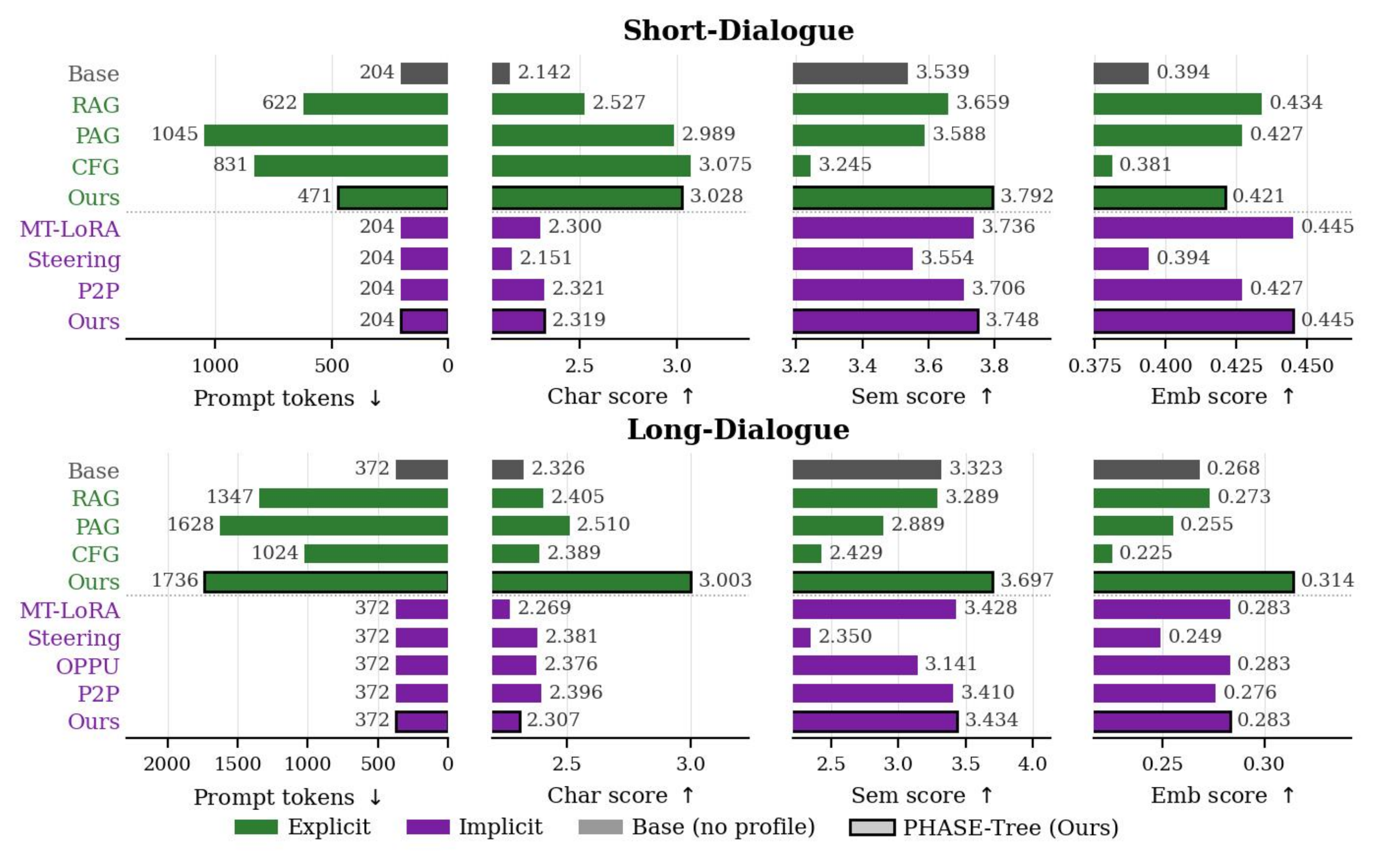}
  \caption{Prompt cost vs.\ quality by horizon. Per row: mean prompt tokens (left; longer bar = higher cost) and average Char, Sem, and Emb (right). Dotted lines separate textual-provision (green) from parametric-adaptation (purple) methods.}
  \label{fig:token-efficiency}
  \end{figure}

\paragraph{Token Efficiency.}
The two paradigms occupy complementary cost--quality operating points (Figure~\ref{fig:token-efficiency}; full breakdown in Appendix~\ref{app:token-cost}). Parametric adaptation eliminates all profile tokens from the prompt (matching the Base cost of 204 short / 372 long tokens), whereas textual provision requires carrying the profile explicitly: 471 tokens on short dialogues (cheaper than RAG 622, CFG 831, PAG 1045) and 1736 on long dialogues (larger than RAG 1347, due to accumulated evolution history) but yields the best long-dialogue Sem and Emb in our comparison. We therefore treat parametric adaptation as a token-efficient deployment variant of the same underlying state.

\section{Discussion and Conclusion}
\label{sec:conclusion}

We address stale-state failure in long-horizon role-playing. PHASE-Tree separates an immutable identity root from editable persona, session, and moment fields. The same flattened state drives generation either as explicit prompt text or as implicit adapter weights. We introduce LongEvoRoleBench, unifying eight existing corpora under one next-utterance protocol. Its long-dialogue split tests cross-episode evolution; its short-dialogue split tests within-scene state tracking.

Under explicit textual provision, our method ranks first on 21 of 24 internal cells and all 12 long-dialogue external-comparison cells. It improves long-dialogue character score, semantic score, and embedding similarity by 19.7\%, 12.4\%, and 15.1\% over the strongest textual baseline. Under implicit parametric adaptation, it places in the top two on 18 of 24 cells. The gap between paradigms points to profile-to-LoRA compression as the bottleneck, not the tree representation itself. The long-dialogue Sem finding persists across judge models and generation backbones, aligns with blinded human ratings, and is supported by the audit of accepted state updates.

This work provides a validated state representation, benchmark, and textual pipeline for evolution-aware role-playing. Natural extensions include more expressive parametric encoders and learned gating for cross-episode updates. Together, these contributions establish evolution-aware role-playing as a first-class subtask alongside persona preservation and memory recall.

\clearpage

\bibliographystyle{plainnat}
\bibliography{main}

\clearpage

\appendix

\section{Extended Related Work}
\label{app:related-work-extended}

This appendix provides a fuller discussion of the four related research lines and their connections to PHASE-Tree.

\subsection{Role-Playing Dialogue and Persona Fidelity}

LLM role-playing has been studied through prompting, data construction, evaluation, and parameter adaptation. Prompt-based and profile-based systems inject character descriptions, memories, or demonstrations into the context window \citep{characterllm2023,rolellm2024,chatharuhi2023}, while evaluation-oriented work measures whether agents preserve personality fidelity and established-role behavior \citep{incharacter2024,coser2025}. Training-based approaches instead adapt model parameters, either through per-character/per-user PEFT (OPPU; \citealp{oppu2024}) or profile-conditioned adapter generation (P2P; \citealp{tan2025p2p}). These methods provide strong baselines for role fidelity, but their conditioning signal is typically a static profile, a retrieved context, or an adapter tied to a fixed description. We instead study whether and how the conditioning signal itself should evolve with the narrative, and how to evaluate that evolution.

Retrieval-augmented role-playing is complementary to our setting. Retrieving previous utterances or character memories can enrich local context, but retrieval does not by itself specify which parts of a character should remain stable, which should adapt within a session, and which should change only after accumulated evidence. Our focus is therefore not a replacement for retrieval, but a structured state representation and update policy compatible with either retrieved evidence or explicit dialogue histories.

\subsection{Long-Horizon Persona Dynamics and Evaluation}

Recent long-horizon dialogue systems increasingly model profiles and memories as dynamic rather than fixed. DuLeMon studies long-term persona memory in open-domain dialogue \citep{duelemon2022}, DEEPER refines user personas from streaming behavior \citep{deeper2025}, LD-Agent combines event extraction, persona inference, and response generation for long-term dialogue \citep{ldagent2025}, and CharacterGPT reconstructs character personas from chapter-wise narrative summaries \citep{charactergpt2025}. Other contemporaneous systems explore user-centric memory trees, adaptive memory stores, multimodal persona memories, or personality vectors with temporal updates \citep{insideout2026,adamem2026,personavlm2026}, while state-machine approaches model personality transitions explicitly \citep{dynamicpersonality2026}. On the evaluation side, LoCoMo measures very long-term conversational memory \citep{locomo2024}, HorizonBench evaluates evolving user preferences \citep{horizonbench2026}, Persistent Personas measures fidelity degradation in extended interactions \citep{persistentpersonas2026}, PERSONAMEM benchmarks memory-driven persona tracking \citep{personamem2025}, SPASM addresses persona drift in multi-turn simulation \citep{spasm2026}, and MDRP diagnoses memory utilization for role-playing \citep{mdrp2026}.

Because this area is crowded, we keep the PHASE-Tree claim narrow: it is a multi-timescale character state with anchored identity information at the root, editable fields below it, and resistance, evidence, and cooldown gates for deciding when those fields change. We evaluate it through a unified protocol over eight existing role-playing corpora. We do not position the system as introducing dynamic personas or persona evolution per se. Unlike user-centric memory or preference benchmarks, PHASE-Tree evaluates fictional role characters whose relationships, stance, and affect evolve with narrative events. Unlike chapter-level reconstruction or coarse state transitions, PHASE-Tree operates at the field level. PDD also studies dynamic persona conditioning by estimating context-dependent persona-attribute importance at decoding time \citep{pdd2026}; PHASE-Tree instead maintains a persistent evolving state and evaluates whether that state better supports role-playing generation.

\subsection{Structured and Psychology-Grounded Character Modeling}

PHASE-Tree is also related to work that gives character or persona modeling a richer structure. Chameleon argues that state can explain more variation than trait and that LLMs are often state-blind \citep{chameleon2026}; SCOPE motivates socially grounded persona construction \citep{scope2026}; BIG5-CHAT and P-React use psychological traits to shape generation or reactions \citep{big5chat2025,preact2025}; and InCharacter evaluates role-playing agents through psychological interviews \citep{incharacter2024}. These works show that psychology-grounded persona modeling is not new by itself. Rather than adopting psychology as a theoretical commitment, we use it operationally: the time-scale distinction directly determines which fields are stable, which accumulate within a scene, and which refresh at the moment level.

Several recent works are close to ours on structure. Inside Out maintains a user-centric PersonaTree with explicit memory operations \citep{insideout2026}, CFSM codifies character behavior as finite-state machines \citep{cfsm2026}, the Identity-Driven Hierarchical framework \citep{identityhierarchical2025} adds a hierarchical identity backbone to role-playing agents, TaciTree organizes multi-session personalized conversation through a hierarchical tree \citep{tacitree2025}, and Jun et al.\ analyze character-profile axes and bottlenecks in role-playing agents \citep{profileaxes2026}. The closest distinction is in what each node carries: PHASE-Tree separates an anchored identity root from three evolving time-scale strata, treats every editable field as an update target, and evaluates the resulting tree as a generation condition under both short-dialogue and long-dialogue settings.

\subsection{Parametric Personalization and Dynamic Adapters}

Parameter-side personalization has a separate lineage. P2P maps profiles to LoRA adapters through a hypernetwork \citep{tan2025p2p}; HyperLoRA, Generative Adapter, and SHINE show that hypernetworks can generate low-rank adapters for task or context adaptation \citep{hyperlora2024,generativeadapter2025,shine2026}. In role-playing, Neeko uses dynamic LoRA for multi-character agents \citep{neeko2024}, HyCoRA generates role-specific LoRA from character embeddings \citep{hycora2026}, PALACE combines persona-aware prompting with VAE-LoRA for multi-session personalized dialogue \citep{palace2025}, and P-React synthesizes topic-adaptive personality reactions through specialized LoRA experts \citep{preact2025}. PERSONA controls personality at inference time through activation-vector algebra rather than LoRA \citep{persona2026}. Profile-to-LoRA, dynamic LoRA, and activation-space steering are therefore well-established directions on the parameter side, each addressed by multiple concurrent and prior systems.

In this paper, implicit parametric adaptation is a complementary, token-efficient deployment variant of PHASE-Tree rather than the main contribution. The same flattened state is encoded by a profile-to-LoRA hypernetwork, removing the profile text from the dialogue prompt. The primary validated path is explicit textual provision, where the evaluation protocol directly measures whether the tree-mediated state yields stronger role-playing responses.

\section{PHASE-Tree Schema}
\label{app:schema}

Table~\ref{tab:schema} gives the full PHASE-Tree schema, including all fields under each block and the corresponding psychological grounding.

\begin{table}[htbp]
  \centering
  \small
  \setlength{\tabcolsep}{8pt}
  \renewcommand{\arraystretch}{1.15}
  \caption{PHASE-Tree schema. Identity anchors fixed name and gender plus append-only backstory; Persona, Session, and Moment are the three time-scale strata used for character-state tracking and evolution.}
  \label{tab:schema}
  \begin{tabular}{@{}c p{0.46\textwidth} p{0.30\textwidth}@{}}
  \toprule
  \textbf{Block} & \multicolumn{1}{c}{\textbf{Fields (with types)}} & \multicolumn{1}{c}{\textbf{Psych.\ Grounding}} \\
  \midrule
  Identity & name (str), gender (str), backstory (str, $\leq$30 EN words / 40 ZH chars initially) & Name and gender fixed; backstory append-only \\
  Persona  & personality, speaking style, behavioral tendencies, hobbies, occupation, demographics (free-text str); relationships (str, ``role is name'' pattern) & Dispositional traits \citep{mcadams1995what,mcadams2006new} \\
  Session  & learned information, commitments, stance changes (\texttt{list[str]}); attitude shifts (\texttt{dict[name$\to$str]}) & Characteristic adaptations \citep{mcadams1995what} \\
  Moment   & emotion (str label), emotion intensity (\texttt{int} 1--10), scene context (str or \texttt{null}) & State affect \citep{spielberger1983manual,rosenberg1998levels} \\
  \bottomrule
  \end{tabular}
\end{table}

\section{Evolution Hyperparameter Details}
\label{app:evolution-params}

This appendix reports the resistance-tier thresholds, cooldown durations, evidence-archive policies, and merge-type rules used in cross-episode persona evolution. All values are the implementation defaults applied uniformly across the four long-dialogue corpora; they were chosen so that core-tier updates require evidence spanning roughly two-thirds of a typical television season of a typical television corpus.

\paragraph{Resistance-tier thresholds.}
Table~\ref{tab:resistance-thresholds} lists the per-tier thresholds enforced by the validator. $\tau^{\mathrm{ep}}_{r}$ is the minimum number of distinct episodes that must contribute evidence, $\tau^{\mathrm{high}}_{r}$ is the minimum number of high-significance entries among that evidence, and $\tau^{\mathrm{cd}}_{r}$ is the field-level cooldown in episodes. The rightmost column summarizes the evidence track: Track~A (recent active archive) for one-off factual changes, or Track~B (lifetime high-significance archive) for slow-moving pattern detection. A high-significance event may serve as evidence for both tracks simultaneously.

\begin{table}[htbp]
  \centering
  \small
  \setlength{\tabcolsep}{10pt}
  \renewcommand{\arraystretch}{1.15}
  \caption{Resistance-tier thresholds for cross-episode field updates.}
  \label{tab:resistance-thresholds}
  \begin{tabular}{@{}lccc>{\raggedright\arraybackslash}p{5.2cm}@{}}
  \toprule
  \textbf{Tier} & $\tau^{\mathrm{ep}}$ & $\tau^{\mathrm{high}}$ & $\tau^{\mathrm{cd}}$ & \textbf{Evidence / extra rule} \\
  \midrule
  low      & 1  & 0 & 2  & Track~A; 1 high or 2 medium \\
  moderate & 3  & 0 & 3  & Track~B; consistent direction \\
  core     & 16 & 6 & 16 & Track~B; incremental preferred \\
  \bottomrule
  \end{tabular}
\end{table}
The last column states the evidence track and LLM-side guidance; only the three $\tau$ columns are enforced as hard rejection rules. For the low tier, the $\tau^{\mathrm{ep}}$ test is bypassed when the disjunctive rule in the last column is met (one high-significance entry or two medium-significance entries).

\paragraph{Field-to-tier assignment.}
The mapping from persona fields to resistance tiers is fixed at initial tree extraction and is identical for every character: the \texttt{core} tier covers \{\texttt{speaking\_style}, \texttt{personality}\}; the \texttt{moderate} tier covers \{\texttt{behavioral\_tendencies}\}; and the \texttt{low} tier covers \{\texttt{hobbies}, \texttt{relationships}, \texttt{occupation}, \texttt{demographics}\}. Name and gender are immutable. Backstory has no resistance tier and may receive an append-only event summary when a persona update is accepted; factual changes such as age, location, and life stage are written to \texttt{demographics}.

\paragraph{Archive policy.}
Session entries are retained according to their significance label: high-significance entries never expire, whereas medium-significance entries are marked \texttt{expired} once they have been carried for more than $8$ episodes without being consumed. When the validator queries the archive, it shows the LLM at most the $20$ most recent medium-significance entries together with all surviving high-significance entries, keeping the prompt bounded while preserving the lifetime pattern signal needed for moderate- and core-tier judgments.

\paragraph{Merge operations.}
Each accepted field update is tagged with one of two merge types. \textsc{Incremental} merges add or refine content while preserving the previous value (covering both list-append and in-place refinement of free-text fields), and \textsc{replacement} merges substitute the previous value entirely. \textsc{Replacement} is reserved for explicit contradictions (e.g., ``boyfriend is Paolo'' $\to$ ``ex-boyfriend is Paolo'' after an on-screen breakup), and the validator further requires that any \textsc{incremental} merge preserve at least $80\%$ of the previous value's length and at least $50\%$ of its distinctive words.

\section{Long-Dialogue Evolution Pipeline}
\label{app:pipeline}

For long-dialogue corpora, cross-episode persona evolution follows the operational pipeline below (Stages~A--C). This decomposition differs from the three conceptual steps in the main text (evidence accumulation, resistance-gated judgment, incremental field update): Stage~C here denotes only the deterministic post-update patches, not the merge step itself. Stage~A is a separate preprocessing pass; Stage~B runs inside the episode-wise orchestrator after each episode; Stage~C applies six patches (plus one optional audit) once Stage~B has finished over all episodes. Each step in Stages~B and~C writes a timestamped persona-snapshot backup and a per-step log, so the full pipeline is reversible.

\paragraph{Stage A: Evidence accumulation.}
Before the episode-wise orchestrator starts, an LLM session-extraction pass reads each scene in episode order and, from each main character's perspective, writes a session-archive entry labelled \texttt{high}, \texttt{medium}, or \texttt{low}. Only high- and medium-significance entries are persisted to the per-character archive used by Stage~B; low-significance entries are discarded. Status-changing events (relationship transitions, occupation changes, major life revelations, verbal turning-point commitments) are required to be labelled \texttt{high} so that they can trigger persona updates downstream.

\paragraph{Stage B: Resistance-gated field update.}
After each episode, the validator builds, for each character, the active session archive (recent unconsumed entries plus the lifetime high-significance arc) and asks an LLM to propose per-field updates conditioned on this archive. Each proposed update is then checked against the tier thresholds described above (episode count, high-significance count, field cooldown), the scene-ID whitelist (every cited \texttt{consumed\_session\_id} must come from the active archive), and an incremental safety check on non-replacement merges; proposals that fail any check are rejected and the field is left unchanged. Accepted updates are written to a new persona snapshot version, an optional event summary is appended to backstory, and consumed archive entries are marked so they are not double-cited later.

\paragraph{Stage C: Post-update patches.}
Six deterministic patches then run in batch on the resulting snapshots, in a fixed order, plus one optional LLM audit that is disabled by default:
\begin{enumerate}[leftmargin=*]
    \item \textbf{Stale-romantic decay.} Demote ``current'' romantic partners (boyfriend / girlfriend / partner / etc.) that no longer have recent fresh evidence in the archive.
    \item \textbf{Inter-main reciprocity repair.} Demote short-lived, unreciprocated romantic claims on a main character, and propagate sustained reciprocal roles to the interacting main when one side asserts a current couple-tier relationship for several consecutive episodes without a matching entry on the other.
    \item \textbf{Legacy relationship normalization.} Normalize bare-name and plural-\texttt{are} legacy formatting so that every entry follows the canonical ``role is name'' pattern.
    \item \textbf{Inverse-pair alignment.} A two-pass step that first demotes premature partner roles (e.g., ``husband'' before an on-screen wedding) and then aligns the partner tier between bidirectional entries.
    \item \textbf{Continuity forward-fill.} Fill 1-to-$N$ episode regression gaps in main-couple relationships when both sides match before and after the gap, no breakup evidence appears in the archive, and the gap is caused by missing evidence in transitional episodes.
    \item \textbf{Core-trait audit} (optional, off by default). A periodic, descriptor-level LLM audit of \texttt{personality} and \texttt{speaking\_style} at coarse checkpoints (default: every 24 episodes plus the finale); at most one descriptor may change per audit, each supported by at least two high-significance session IDs from the lifetime archive.
\end{enumerate}

\paragraph{Human validation of field updates.}
Three annotators evaluated a stratified 30\% sample of evidence-qualified persona snapshots from the four long-dialogue corpora (130 snapshots and 152 accepted field updates). Each item presented the previous field value, the extracted evidence and update rationale, and the updated value. Annotators focused on the substantive change between the two values and assessed how well it reflected the character evidence available through the target episode. Direct statements and actions received the greatest weight, while stable personality or behavioral changes were assessed from recurring patterns or a clear turning point. They assigned one of three labels: supported, partially supported, or unsupported, distinguishing updates that were fully grounded in the evidence from those that captured the general direction but were indirect, incomplete, or overly broad. Table~\ref{tab:state-validation} gives the majority label under an evidence-support rubric and a stricter semantic rubric that also considers how precisely the updated value expresses the evidence.

\begin{table}[!ht]
\centering
\small
\setlength{\tabcolsep}{8pt}
\renewcommand{\arraystretch}{1.15}
\caption{Majority-consensus labels on the 152 accepted field updates (three annotators).}
\label{tab:state-validation}
\begin{tabular}{@{}lcccc@{}}
\toprule
\textbf{Rubric} & \textbf{Supported} & \textbf{Partially} & \textbf{Unsupported} & \textbf{Fleiss $\kappa$} \\
\midrule
Evidence support     & 77.6\% & 22.4\% & 0.0\% & 0.47 \\
Semantic (stricter)  & 65.1\% & 34.9\% & 0.0\% & 0.43 \\
\bottomrule
\end{tabular}
\end{table}

Under both rubrics, every sampled update is supported or partially supported; none is labelled unsupported. Inter-annotator agreement is moderate (Fleiss $\kappa=0.43$--$0.47$). The difference between the two rubrics appears mainly in whether an update is fully or partially supported, rather than whether it lacks narrative support.

\section{Pipeline Prompts}
\label{app:prompts}

This appendix gives the four prompt specifications that drive tree construction, intra-episode session/moment extraction, and the two cross-episode evolution stages. All use a frozen GPT-4.1 endpoint at \texttt{temperature=0}; their source files are under \texttt{src/tree\_pipeline/}. The boxes retain the operative system rules and user-message slot structures while omitting only redundant demonstrations; \texttt{\{braced\}} slots are filled at runtime. The long-dialogue prompts replace \texttt{\{show\}} with the corpus name.

\subsection{Initial Tree Extraction}
\label{app:prompts:p1}

Tree-construction prompt, defined as \texttt{SYSTEM\_PROMPT} in \texttt{profiles\_to\_trees.py}.

\begin{promptbox}
You are a character attribute extraction specialist. Your task is to read a fictional character's raw profile data (which may contain irregular, redundant, or heterogeneous fields) and extract it into a standardized attribute tree JSON.

## CRITICAL RULES -- READ BEFORE ANYTHING ELSE

1. **USE ONLY THE PROVIDED DATA.** Extract information EXCLUSIVELY from the raw profile JSON. Do NOT add, infer, or supplement from your own training knowledge.
2. **NULL FOR MISSING FIELDS.** If the raw profile contains no information for a field, set its value to null. Do NOT fabricate; null is always better than a hallucination.
3. **NO CHARACTER NAME IN VALUES.** The name lives in identity.name; do NOT mention it inside any other value field.
4. **SEMANTIC ANALYSIS -- DO NOT BLINDLY COPY SOURCE FIELDS.** Source profiles may group heterogeneous information under one field; analyze each fact and route it to the single target field whose definition it best matches. A single source field may need to be SPLIT across multiple target fields.
5. **STRICT NO-DUPLICATION -- ZERO TOLERANCE.** Each fact appears in exactly ONE field. Job titles only in occupation; personality traits only in personality; relationship details only in relationships; demographics only in demographics.
6. **SKIP TRIVIALLY OBVIOUS INFO.** For a human in a real-world setting, do NOT write "human" in demographics. Only include species/race when non-human or otherwise distinctive.
7. **OUTPUT LANGUAGE.** Write all value strings in the same language as the input raw profile data; field keys are always English.
8. **PARAPHRASING TOLERANCE.** Light reorganization only (up to half a sentence of connective phrasing or omission). No heavy rewriting or embellishment beyond the source.
9. **NO COLONS IN VALUES -- USE NATURAL LANGUAGE.** All value strings must be flowing natural language. No "key: value" formatting inside any value string; use connectors such as "is" / "includes".

## Attribute Tree Structure

The tree has four layers: identity, persona, session, moment. You only fill identity and persona; session and moment use defaults.

### identity layer (fixed name and gender; append-only backstory)
- **name**: The character's formal full name only. No nicknames or aliases.
- **gender**: Gender, or null if not explicitly stated and not clearly inferable.
- **backstory**: A VERY BRIEF summary of the 1-2 most pivotal life turning points. **HARD LIMIT: 1 sentence, max 40 ZH chars / 30 EN words.** Exclude occupation/job titles, personality traits, relationship names, demographic facts.

### persona layer (slowly-changing traits)

Each field has a text value and a resistance level.

- **speaking_style** (resistance: core): A description of HOW the character speaks, NOT the quotes themselves. Summarize speech patterns in descriptive terms; null if no speech-related information exists.
- **personality** (resistance: core): Innate character traits, temperament, moral outlook, and emotional tendencies. Hobbies and work habits do NOT belong here.
- **behavioral_tendencies** (resistance: moderate): Recurring action patterns, work style, social conduct, talents, specific skills, combat techniques. Life events / plot points / job titles / inferred abilities / power levels do NOT belong here.
- **hobbies** (resistance: low): Interests, hobbies, specific likes and dislikes. Innate traits and professional skills do NOT belong here.
- **relationships** (resistance: low): Key interpersonal relationships ONLY. Each item must STRICTLY follow the pattern "ROLE is NAME" -- no colons, no events, no emotional descriptions, no narrative context.
- **occupation** (resistance: low): Job title, professional role, social role.
- **demographics** (resistance: low): Age, height, species (non-human only), residence, education, etc., in natural language without colons. Exclude gender (already in identity.gender), name, occupation, relationships.

## resistance assignment (fixed)
- core: speaking_style, personality
- moderate: behavioral_tendencies
- low: hobbies, relationships, occupation, demographics

## Pre-output self-check (MANDATORY)

Before outputting, verify each field:
1. speaking_style is a DESCRIPTION of how they speak, not raw quotes. If it contains catchphrases verbatim, REWRITE.
2. behavioral_tendencies describes HABITUAL patterns or skills. Remove any life event, plot point, power level, cultivation rank, job title, or inferred ability.
3. relationships items follow the strict "ROLE is NAME" pattern. Remove narrative verbs and emotional descriptions; if an item is purely an event with no role+name, DELETE it.
4. backstory is within the 30 EN words / 40 ZH chars limit and contains no job title or occupation keyword.
5. No fact is duplicated across two or more fields; keep it in the most specific field.
6. No value string contains a colon used as a key-value separator.
7. demographics / relationships value is set to null if ALL sub-items are null/unknown.

## Output format

Output ONLY a valid JSON object. No extra text, explanations, or markdown fences.

{
  "identity": {
    "name": "...",
    "gender": "...",
    "backstory": "..." or null
  },
  "persona": {
    "speaking_style":        {"value": "..." or null, "resistance": "core"},
    "personality":           {"value": "..." or null, "resistance": "core"},
    "behavioral_tendencies": {"value": "..." or null, "resistance": "moderate"},
    "hobbies":               {"value": "..." or null, "resistance": "low"},
    "relationships":         {"value": "..." or null, "resistance": "low"},
    "occupation":            {"value": "..." or null, "resistance": "low"},
    "demographics":          {"value": "..." or null, "resistance": "low"}
  }
}
\end{promptbox}

User-message template from \texttt{build\_user\_prompt}:
\begin{promptbox}
Below is the raw profile data for the character "{char_key}" in JSON format.
Extract and organize it into a standardized attribute tree following the system instructions.

REMINDER:
- Use ONLY the data below; do NOT use your own knowledge about this character.
- Analyze each fact and place it in the SINGLE best-matching field.
- A source field may need to be SPLIT across target fields.
- NEVER duplicate facts across fields.
- Work habits and social conduct go in behavioral_tendencies.
- Hobbies, likes, and dislikes go in hobbies.
- Job titles go ONLY in occupation.
- Do NOT include the character's name in any value string.

{raw_profile_json}
\end{promptbox}

\subsection{Intra-Episode Session and Moment Extraction}
\label{app:prompts:p2}

Session- and moment-extraction prompt, defined as \texttt{SYSTEM\_PROMPT} in \texttt{update\_session\_moment.py} and re-imported by \texttt{extract\_long\_term\_moments.py}, so short- and long-dialogue tracks share it verbatim.

\begin{promptbox}
You are a character state analyst. Given a character's persona profile and a dialogue history, your task is to analyze WHAT CHANGED for the character during this conversation and output updated session and moment fields.

## CRITICAL RULES

1. **PERSONA IS FROZEN.** Do NOT modify or comment on the persona layer. Your job is ONLY to fill session and moment based on the dialogue.
2. **ANALYZE FROM THE CHARACTER'S PERSPECTIVE.** The "role" field tells you which character you are analyzing. Only track what THIS character learned, felt, or committed to -- not other speakers.
3. **USE ONLY DIALOGUE EVIDENCE.** Every field you fill must be grounded in something explicitly said or clearly implied in the dialogue. Do NOT fabricate.
4. **BE CONCISE.** Each learned_info item is one short sentence; attitude_shifts values are brief descriptions; emotion is a single word or short phrase.
5. **NULL/EMPTY FOR MISSING INFO.** If the dialogue is too short or simple to extract meaningful changes, return empty lists/objects. Do NOT invent content.
6. **OUTPUT LANGUAGE.** Write all values in the SAME language as the persona profile; the required language is specified explicitly in the user message.
7. **THIRD-PERSON PERSPECTIVE.** You MUST write ALL session and moment content in THIRD PERSON. Never use first-person pronouns to refer to the character being analyzed; use the character's name or third-person pronouns instead.

## Fields to fill

### session (cumulative within this dialogue)
- **learned_info** (list of strings): New information the character learned during this conversation.
- **attitude_shifts** (object, key=person, value=description): How the character's attitude toward specific people changed.
- **commitments** (list of strings): Promises or decisions the character made.
- **stance_changes** (list of strings): Shifts in the character's position or viewpoint.

### moment (state at the END of this dialogue)
- **emotion** (string): The character's dominant emotion at the conversation's end. Use a concise label.
- **emotion_intensity** (int 1-10): How strong the emotion is.
- **scene_context** (string or null): Brief description of the scene/situation.

## Output format

Output ONLY a valid JSON object with exactly two keys: "session" and "moment". No extra text, explanations, or markdown fences.

{
  "session": {
    "learned_info":    [...],
    "attitude_shifts": {...},
    "commitments":     [...],
    "stance_changes":  [...]
  },
  "moment": {
    "emotion":           "...",
    "emotion_intensity": N,
    "scene_context":     "..." or null
  }
}
\end{promptbox}

User-message template from \texttt{build\_user\_prompt}:
\begin{promptbox}
## Output language requirement
{language_instruction}

## Character being analyzed: {role}

## Character persona (frozen, for reference only):
{persona_summary}

## Dialogue history (analyze this to update session & moment):
{dialogue_input}
\end{promptbox}

\subsection{Evidence Accumulation (Stage A)}
\label{app:prompts:p3}

Per-scene significance-labelling prompt, defined as \texttt{SYSTEM\_PROMPT\_TEMPLATE} in \texttt{extract\_evolution\_sessions.py}.

\begin{promptbox}
You are a character development analyst for the TV show "{show}". Given a scene's full dialogue and a specific character's current persona, evaluate what happened in this scene FROM THAT CHARACTER'S PERSPECTIVE and assess its significance for their long-term character development.

## CRITICAL RULES

1. Analyze ONLY from the specified character's perspective.
2. Use ONLY evidence from the provided dialogue. Do NOT use external knowledge about the show's future plot.
3. Write in THIRD PERSON. Never use "I", "my", "me".
4. ALL output must be in English.
5. If the character is barely involved or the scene has no meaningful impact on them, set significance to "low".

## Significance levels

- **high**: ALWAYS use "high" when the scene EXPLICITLY shows any of:
  * a relationship STATUS change (breaking up, getting together, engagement, marriage, divorce, learning of a pregnancy/birth, gaining/losing a child, losing a pet, reconciliation)
  * an occupation/career change (hired, fired, quit, promoted, audition won)
  * a major life revelation (parent's affair, learning a hidden truth, coming out, moving)
  * a turning-point decision the character verbally commits to
- **medium**: Notable interactions that reveal or slightly shift character traits (e.g., a meaningful conversation, minor conflict, learning something new) without producing a status change.
- **low**: The character is barely present, only makes small talk, or nothing in the scene has any meaningful impact on their development.

IMPORTANT: When in doubt between high and medium for a relationship/job/family status change, choose **high**. Status transitions MUST be high so they can trigger a persona update.

## Output -- valid JSON only, no markdown fences

{
  "summary":         "One sentence describing what this scene meant for the character (third person)",
  "significance":    "high/medium/low",
  "affected_fields": ["persona fields potentially affected, e.g. personality, relationships, occupation, behavioral_tendencies"]
}

If significance is "low", summary should be brief and affected_fields can be an empty list.
\end{promptbox}

User-message template:
\begin{promptbox}
## Character being analyzed: {character}

## Character's current persona (for reference):
{persona_summary}

## Scene dialogue:
{scene_dialogue}
\end{promptbox}

\subsection{Resistance-Gated Judgment (Stage B)}
\label{app:prompts:p4}

Per-episode field-update prompt, defined as \texttt{\_SYSTEM\_PROMPT\_TEMPLATE} in \texttt{evolve\_persona.py}. This is the longest of the four prompts; we keep its rule headers verbatim and elide only the in-prompt narrative examples.

\begin{promptbox}
You are a character psychologist specialising in long-term personality development for the TV show "{show}". Given a character's current persona and their recent experiences (session archive), decide whether the accumulated experiences warrant updating any persona field.

## Field semantics -- IMPORTANT

Use each persona field for its intended purpose. Do NOT cross-pollute fields:
- **occupation** = job / profession / career. NEVER put pets, hobbies, family roles or romantic statuses here.
- **relationships** = social bonds (family, friends, romantic partners). List each person at most ONCE with their CURRENT role. **Direction matters.** The format is "<the OTHER person's role to YOU> is <Name>"; the role describes how the named person relates to the persona character, not vice versa.
- **hobbies** = leisure activities the character enjoys.
- **demographics** = age, location, life stage.
- **personality** = core inner traits (adjectives describing the person).
- **speaking_style** = how they talk.
- **behavioral_tendencies** = recurring behaviour patterns.

When an event affects multiple fields, update all relevant fields in the same decision so the persona stays internally consistent.

## Two-track evidence views

You will see TWO archive sections in the user message:

- **Track A -- RECENT EVIDENCE** (active high-sig + recent medium events since the last persona update). Primary view for **low-resistance factual fields** (relationships, occupation, demographics, hobbies). These fields update on specific recent events.
- **Track B -- LIFETIME PATTERN ARCHIVE** (every high-significance event ever produced for this character up to the current episode, in chronological order). Primary view for **moderate / core fields** (behavioral_tendencies, personality, speaking_style). A high-significance event is evidence for BOTH a one-off factual update AND for slow-moving pattern detection -- its appearance here does NOT mean it is "reusable" for low-field updates.

## Three-tier resistance system

- **low** (occupation, relationships, demographics, hobbies): Use Track A. Update when there is at least 1 high-significance event OR 2 medium events providing CLEAR FACTUAL EVIDENCE. Evidence must describe an ACTUAL EVENT, not an intention or plan.
- **moderate** (behavioral_tendencies): Use Track B. Update when consistent evidence spans **at least 3 different episodes** in Track B pointing in the same direction.
- **core** (personality, speaking_style): Use Track B. Update when the lifetime arc shows CONSISTENT directional drift across **>= 16 distinct episodes** with **>= 6 high-significance events** all reinforcing the same refinement. **Prefer INCREMENTAL refinements** (adding nuance, replacing one outdated descriptor) over wholesale REPLACEMENT. Do NOT update core on a single dramatic episode, no matter how high-significance.

## Core-field re-examination trigger (IMPORTANT)

Each persona field carries a times_updated counter and a last_updated tag. A "DRIFT TRIGGER" entry in the user message means the behavioural surface (behavioral_tendencies) has updated several times in a consistent direction while the underlying deep trait has not been revisited.

When such a trigger is present AND the core-tier threshold is MET, you **SHOULD explicitly check** whether any single descriptor in the current deep-field value has been **directly contradicted** by the accumulated behavioural drift in Track B.
- If yes, propose an INCREMENTAL refinement: keep all still-accurate descriptors verbatim, replace ONE outdated descriptor or append a small qualifier nuance. Cite >= 6 scene_ids drawn from >= 16 distinct episodes.
- If no, keep the deep field unchanged. Drift in surface behaviour does NOT automatically imply drift in deep traits.

This trigger is NOT a mandate to update -- it is a structured nudge to examine. Spurious or weakly-supported core changes will be rejected.

## Citing evidence in consumed_session_ids

For **low** fields, cite scene IDs from Track A only.
For **moderate** / **core** fields, cite scene IDs from Track B; re-citation across tiers is permitted (one significant event can serve as evidence for both an immediate factual change and the long-arc pattern).
Always cite at least 3 IDs for moderate updates and at least 6 IDs for core updates, drawn from at least 3 / 16 distinct episodes respectively.

## Merge strategy: CONFLICT-BASED

For each field you decide to update, classify the relationship between the new evidence and the existing value:
- **Case 1 -- NO CONFLICT (additive)**: New information is compatible with all existing facts. Use **incremental** merge: KEEP ALL existing content, append or weave in the new info.
- **Case 2 -- CONFLICT (mutually exclusive)**: New information makes an existing specific item logically untrue. Use **replacement** merge: replace ONLY the conflicting item, keep everything else verbatim.

## CRITICAL -- Distinguish what the evidence ACTUALLY shows

Session summaries describe events from the focal character's perspective. Do NOT over-interpret:
- "X reveals/expresses feelings for Y" does NOT mean X confessed to Y directly. The disclosure is often to a THIRD PARTY. Update X-Y's relationship ONLY if the evidence EXPLICITLY shows mutual romantic interaction.
- "X impersonates Y in a conversation with Z" does NOT make X and Z a couple.
- "X helps Y break up with Z" does NOT make X and Z a couple.
- "X comforts Y after a breakup / vulnerable moment" does NOT establish a new romantic relationship between X and Y.
- A character's INNER FEELINGS (longing, jealousy, attraction) do NOT count as a relationship status change. The status only changes when ACTIONS occur (mutual kiss, date, declaration, breakup, etc.).

## ABSOLUTELY CRITICAL -- NO INFERENCE FROM ABSENCE

**NEVER remove or downgrade an existing fact just because the recent sessions don't mention it.** Absence of mention is NOT evidence of contradiction. A fact stays in the persona until the dialogue EXPLICITLY shows it changed. NEVER delete information that is not contradicted by the new event.

## CRITICAL -- Romantic-status downgrades require an EXPLICIT breakup

Changing "<role> is X" to "ex-<role> is X" is a **status change**. It is allowed ONLY when the cited sessions explicitly describe an actual breakup, divorce, calling-it-off, or split -- not arguments, distance, jealousy, fights, hurt feelings, or temporary silence.

## DO add new partners and reconciliations as they appear

The "no inference from absence" rule above bars REMOVING facts; it does NOT discourage APPENDING new factual relationship entries. When a session explicitly introduces a new dating partner (even for a short arc) or shows reconciliation with an estranged friend, you SHOULD update relationships to include them. When in doubt, err toward including; the validators will reject genuinely unsupported claims.

## Evidence reporting

For each field change, list the SPECIFIC scene_ids of the sessions that support that particular change. Only cite scene_ids that appear in the provided session archive. Scene-ID format MUST be exactly as shown in the archive (lowercase, e.g. s01_e12_c10); do NOT invent variants.

## Output format -- valid JSON only, no markdown fences

{
  "should_update": true or false,
  "reasoning": "Brief explanation of why this update is or is not warranted",
  "changes": {
    "field_name": {
      "new_value": "complete new value of the field",
      "merge_type": "incremental" or "replacement",
      "consumed_session_ids": ["scene_id_1", "scene_id_2", ...]
    }
  },
  "backstory_addendum": "One sentence to append to identity.backstory" or null
}

If should_update is false, changes must be {} and backstory_addendum must be null. If you can update some fields but not others, include only the fields you can justify; do NOT include weakly-supported changes.
\end{promptbox}

User-message template; the drift-signal block is included only when a trigger fires:
\begin{promptbox}
## Character: {character}

## Current identity
Backstory: {backstory}

## Current Persona (version {version})
{persona_text}

## Lifetime evidence summary (Track B)
- Unique episodes with high-sig events: {unique_episodes}
- Total high-sig events: {high_events}
- Moderate-tier threshold (>=3 eps): {MET | not yet met}
- Core-tier threshold (>=16 eps AND >=6 high): {MET | not yet met}

If a tier's threshold is not yet met, fields at that tier cannot update.

## Deep-field drift signals [optional]
- DRIFT TRIGGER: '{surface}' has been updated {N} time(s) while '{deep}' has stayed unchanged. Examine whether the accumulated drift contradicts a descriptor in '{deep}'.

## Track A -- RECENT EVIDENCE
{K} entries:
{track_a_text}

## Track B -- LIFETIME PATTERN ARCHIVE
{L} entries:
{track_b_text}
\end{promptbox}

\section{Dataset Statistics and Split Construction}
\label{app:datasets}

Table~\ref{tab:datasets} lists the eight source corpora or resources, output languages, main-character counts, and benchmark conversions.

\begin{table}[htbp]
  \centering
  \small
  \setlength{\tabcolsep}{8pt}
  \renewcommand{\arraystretch}{1.1}
  \caption{Dataset statistics and output language. Short-dialogue profiles are static; long-dialogue states may evolve across episodes.}
  \label{tab:datasets}
  \begin{tabular}{@{}c l c c >{\raggedright\arraybackslash}p{6.6cm}@{}}
  \toprule
  \textbf{Type} & \textbf{Dataset} & \textbf{Lang.} & \textbf{\#Chars} & \multicolumn{1}{c}{\textbf{Note}} \\
  \midrule
  \multirow{4}{*}{\shorttag{Short}} & \shorttag{RAIDEN} & ZH & 30 & RPCA benchmark; light conversion \\
   & \shorttag{CharacterEval} & ZH & 77 & RPCA benchmark; light conversion \\
   & \shorttag{SimsConv} & EN & 68 & Simulated conversations; profiles parsed from instructions \\
   & \shorttag{ChatHaruhi} & EN/ZH & 31 & Anime/fiction roles; profiles synthesized from dialogues \\
  \midrule
  \multirow{4}{*}{\longtag{Long}} & \longtag{Friends} & EN & 6 & ConvoKit; 10 seasons \\
   & \longtag{The Office} & EN & 6 & Public transcripts; 9 seasons \\
   & \longtag{Harry Potter} & EN & 6 & HPD/book dialogue \\
   & \longtag{Star Trek} & EN & 6 & Public TNG scripts; 7 seasons \\
  \bottomrule
  \end{tabular}
\end{table}

\paragraph{Short-dialogue corpora.}
RAIDEN and CharacterEval are existing RPCA benchmarks and are used after lightweight conversion into the common profile--context--target fields \citep{raiden2025,charactereval2024}. SimsConv and ChatHaruhi are also existing role-playing resources, but require additional preprocessing to align character profiles, scene contexts, target utterances, and the OOD character-cluster split \citep{simsconv2025,chatharuhi2023}. In all four short-dialogue sets, the model conditions on the given profile (or its ablation/tree variant) and produces an in-character response. There is no cross-episode narrative axis; only intra-episode session and moment tracking applies.

\paragraph{Long-dialogue corpora.}
Friends is processed from the ConvoKit Friends Corpus \citep{convokit2020}; The Office and Star Trek are processed from public episode transcript resources, with per-episode source identifiers retained in the release metadata; Harry Potter dialogue is drawn from HPD, a previously released character-aligned dialogue source \citep{hpd2023}. For all four long-dialogue corpora, we construct episode- or book-indexed next-utterance instances and temporal holdouts rather than adopting an existing benchmark split. Each set follows six main characters whose beliefs, relationships, and affect may shift over a long arc, providing the setting for cross-episode persona evolution.

\paragraph{OOD split construction.}
For short-dialogue data, we embed profile text and cluster characters by similarity. Clusters are ranked by average inter-cluster distance with a size penalty, and a fixed target number of OOD characters is drawn from the highest-ranked clusters; random-test characters are selected to a separate fixed target by cluster-stratified sampling from the remainder. For long-dialogue data, we split temporally: Friends and The Office hold out the last three seasons (8--10 and 7--9); Star Trek holds out TNG seasons 6--7; Harry Potter holds out books 6--7. Earlier seasons or books supply train and random-test episodes.

\section{LLM Judge Details}
\label{app:judge}

We use an LLM-as-Judge for Character Score (Char) and Semantic Score (Sem); the primary ratings are produced by GPT-4.1 under greedy decoding (temperature=0, top\_p=1) through a single fixed endpoint, and the same model is used for PHASE-Tree extraction. Appendix~\ref{app:judge-crossmodel} evaluates the same responses with two additional judges under the same rubric. Each response receives independent 1--5 integer ratings. The verbatim rubric and prompt are released at \texttt{evaluation/persona\_rubric.md}.

\paragraph{Judge constraints.}
(1) Char is based only on the supplied profile text; parametric knowledge of the fictional character is excluded, and traits not stated in the profile are neither rewarded nor penalized. (2) Sem is based only on the dialogue context; the ground-truth response illustrates the kind of conversational moment (humorous, emotional, informational, etc.) and is not treated as the unique correct answer. (3) The two scores are rated independently: a response with poor profile consistency can still be contextually coherent, and vice versa.

\paragraph{Character Score (profile consistency).}
Measures how consistently the response reflects the traits described in the profile (personality, speaking style, emotional tendencies, relationships, behavioral patterns), not authenticity to any real or fictional persona.

\begin{itemize}
    \item 1 (none): Generic, flat, or interchangeable with any identity; no described trait is discernible.
    \item 2 (weak): At most one trait surfaced (e.g., a slightly matching tone); other described traits are absent or contradicted.
    \item 3 (moderate): Two or more described traits recognizable and adapting the tone, but appearing in isolation rather than forming a coherent characterization.
    \item 4 (strong): Multiple described traits integrated coherently across tone, emotional register, and interpersonal dynamics; minor omissions allowed, but no trait is contradicted.
    \item 5 (full): Personality, style, emotional state, and relational dynamics from the profile converge naturally; the response reads as though it could only have been produced under this specific profile.
\end{itemize}

\paragraph{Semantic Score (contextual coherence).}
Measures whether the response is a natural continuation of the dialogue context. Equally valid but different continuations receive Score $\geq 3$; character-style quality is captured by Char and not double-counted here.

\begin{itemize}
    \item 1 (incoherent): Nonsensical, self-contradictory, or unrelated; reads as if inserted from a different conversation.
    \item 2 (marginal): Connects to the scene superficially but misreads the moment (e.g., humorous when the moment is serious, addresses a topic no one raised).
    \item 3 (coherent): Reacts to what was said, matches the expected register, and is a plausible next line, even if it pursues a different angle from the reference.
    \item 4 (aligned): Natural continuation that also addresses the same topic or communicative intent as the reference; specific wording differs but the conversational function overlaps.
    \item 5 (near-equivalent): Same communicative intent, key references, and emotional direction as the reference; wording differs but the conversational effect is interchangeable.
\end{itemize}

\section{Experimental Implementation Details}
\label{app:exp-impl}

The primary experiments use Qwen2.5-7B-Instruct as the frozen backbone; Appendix~\ref{app:backbone} reports additional frozen backbones. Reported model predictions in the PHASE-Tree evaluation pipeline, including textual-provision, parametric-adaptation, and external-baseline runs, are decoded with temperature 0.3, a 256-token maximum, and seed 42. Trees are built with the extraction pipeline described in the method. The parametric-adaptation route warm-starts from a P2P-style profile-to-LoRA checkpoint \citep{tan2025p2p} and is fine-tuned end-to-end; gradients flow through generated LoRA weights only. External baselines use the same backbone where applicable; OPPU requires per-character adapter training.

\section{Internal Ablation Chain}
\label{app:ablation-chain}

All variants share the same backbone, evaluation pipeline, and (for LoRA) hypernetwork; they differ only in how the character condition is prepared.

\begin{table}[htbp]
  \centering
  \small
  \setlength{\tabcolsep}{8pt}
  \renewcommand{\arraystretch}{1.1}
  \caption{Internal ablation chain.}
  \label{tab:ablation-chain}
  \begin{tabular}{@{}c l >{\raggedright\arraybackslash}p{8.6cm}@{}}
  \toprule
  \textbf{ID} & \textbf{Name} & \multicolumn{1}{c}{\textbf{Description}} \\
  \midrule
  Base & Context-Only & Dialogue context only; no profile. \\
  RP   & Raw-Profile  & Unprocessed dataset profile. \\
  NR   & Naive-Rewrite & One-shot LLM rewrite of profile and context; no tree structure. \\
  ST   & Static-Tree  & Construction-time identity and persona; session and moment remain empty; no evolution. \\
  DT   & Dynamic-Tree & Long-dialogue only: evolved persona fields; session and moment omitted. \\
  PT   & PHASE-Tree   & Full pipeline with intra-episode tracking and cross-episode evolution. \\
  \bottomrule
  \end{tabular}
\end{table}

\paragraph{Fairness.}
NR receives only the native profile and context, produces unstructured text, and omits reply directives. Its rewrite is no longer than RP on average, so gains are not due to longer prompts. DT is evaluated only on long-dialogue sets.

\section{External Baselines}
\label{app:external-baselines}

\paragraph{Textual provision.}
RAG follows retrieval-augmented generation \citep{lewis2020rag} and retrieves historical lines into the prompt. PAG follows profile-prompting role-playing systems \citep{characterllm2023,rolellm2024,chatharuhi2023} and augments the prompt with profile-derived text. CFG follows language-model classifier-free guidance \citep{sanchez2024cfg} and amplifies profile influence via dual forward passes at decode time.

\paragraph{Parametric adaptation.}
MT-LoRA trains one shared LoRA adapter \citep{hu2022lora} on all characters' dialogues. Steering injects persona activation vectors at inference \citep{turner2024activation}. OPPU trains a separate per-character adapter \citep{oppu2024}. P2P applies a profile-to-LoRA checkpoint without retraining \citep{tan2025p2p}. Cells marked -- denote unavailable runs; OPPU is evaluated only on the four long-dialogue corpora where per-character adapters were trained.

\section{Statistical Significance and Effect Sizes}
\label{app:stats}

This appendix specifies the question-level paired statistical analysis behind the significance claim attached to Tables~\ref{tab:explicit-full} and~\ref{tab:external-full}. For each named comparison, we treat the two methods' per-question scores on the same (dataset, split) cell as a paired sample, then compute the paired $t$-test $p$-value, the Wilcoxon signed-rank $p$-value, and the paired Cohen's $d$ on the per-question score differences. The released evaluation logs contain the full comparison grid, including the Base, RAG, and P2P reference baselines used for routine reporting; Table~\ref{tab:effect-sizes} below reports the specific pairwise comparisons used by the main-text claims. Per-question samples also yield 95\% confidence intervals through the Student's-$t$ approximation; these are provided in the released logs rather than in the main tables to preserve readability.

\paragraph{Why effect size and not just $p$.}
Per-cell paired sample sizes are large, roughly $10^3$ on short-dialogue test splits and up to about $1.6{\times}10^4$ on long-dialogue test splits. With samples of this size, pooled contrasts can yield small $p$-values even when the macro-averaged effect is small, and individual dataset--split cells need not all satisfy $p<0.001$. We therefore interpret rankings primarily through Cohen's $d$, treating $|d|<0.2$ as practically negligible regardless of how small $p$ is. Table~\ref{tab:effect-sizes} reports macro-averaged Cohen's $d$ for the key comparisons cited in the main text. The Sem effects are consistently above this practical threshold, while the cross-paradigm Ours (under textual provision) vs.\ MT-LoRA (under parametric adaptation) Emb effect is borderline ($d=0.19$), so we treat that embedding gain as small rather than as a large practical effect.

\FloatBarrier
\begin{table}[htbp]
  \centering
  \small
  \setlength{\tabcolsep}{6pt}
  \renewcommand{\arraystretch}{1.1}
  \caption{Long-dialogue macro-averaged paired Cohen's $d$ for key comparisons. Range gives the per-cell $d$ extremes across four datasets $\times$ two splits. Ours denotes PHASE-Tree under textual provision; MT-LoRA is the parametric-adaptation baseline.}
  \label{tab:effect-sizes}
  \begin{tabular}{@{}>{\centering\arraybackslash}p{2.9cm} c c >{\centering\arraybackslash}p{2.6cm}@{}}
  \toprule
  \multicolumn{1}{c}{\textbf{Comparison}} & \textbf{Metric} & \textbf{Macro $d$} & \multicolumn{1}{c}{\textbf{Range}} \\
  \midrule
  PT vs.\ NR          & Char & 0.01 & [$-$0.13, 0.08] \\
  PT vs.\ NR          & Sem  & 0.25 & [0.19, 0.36] \\
  PT vs.\ NR          & Emb  & 0.26 & [0.21, 0.33] \\
  \addlinespace
  PT vs.\ ST          & Char & 0.10 & [$-$0.11, 0.21] \\
  PT vs.\ ST          & Sem  & 0.40 & [0.32, 0.50] \\
  PT vs.\ ST          & Emb  & 0.30 & [0.24, 0.36] \\
  \addlinespace
  Ours vs.\ MT-LoRA   & Char & 0.72 & [0.56, 0.95] \\
  Ours vs.\ MT-LoRA   & Sem  & 0.29 & [0.24, 0.34] \\
  Ours vs.\ MT-LoRA   & Emb  & 0.19 & [0.12, 0.27] \\
  \bottomrule
  \end{tabular}
\end{table}

\paragraph{Scope of the statistical analysis.}
The analysis is question-level paired: within each dataset and split, methods are compared on the same instances, which controls for per-question difficulty. The tests use one generation run per method and therefore quantify question-level variation rather than run-to-run decoding variation. The released evaluation logs include the per-cell differences, confidence intervals, effect sizes, and test statistics.

\section{Judge Robustness and Validity}
\label{app:judge-validity}

We analyze the character and semantic scores along three dimensions: sensitivity to the persona reference, consistency across judge models, and agreement with human ratings.

\subsection{Persona-Reference Sensitivity}
\label{app:judge-ref-ablation}

The default judge condition uses the flattened PHASE-Tree profile as the Character Profile for every method. We also score the same responses using the raw character description---the one to three paragraphs of identity, traits, and stock mannerisms used by RP---as the reference. This changes only the Char and Sem judge inputs; generated responses and embedding scores remain fixed.

\begin{table}[htbp]
    \centering
    \small
    \caption{Judge persona-reference ablation: macro-averaged Char, Sem, and Emb under two judge references, the flattened PHASE-Tree profile (PT-prof) and the raw character description (Raw-prof). $\Delta$ rows give Raw-prof minus PT-prof. \textbf{Bold} = best, \underline{underline} = second best on each score row; DT applies to long-dialogue corpora only (--).}
    \label{tab:judge-ref-ablation}
    \begin{tabular}{ll cccccc}
    \toprule
    \textbf{Metric} & \textbf{Judge ref} & \textbf{Base} & \textbf{RP} & \textbf{NR} & \textbf{ST} & \textbf{DT} & \textbf{PT (Ours)} \\
    \midrule
    \multicolumn{8}{c}{\textit{Short-Dialogue Macro}} \\
    \midrule
    \multirow{3}{*}{Char $\uparrow$}
     & PT-prof       & 2.143 & 2.876 & 3.014 & \textbf{3.068} & -- & \underline{3.028} \\
     & Raw-prof      & 2.026 & \textbf{3.074} & \underline{2.970} & 2.943 & -- & 2.881 \\
     & $\Delta$Char  & $-0.116$ & $+0.198$ & $-0.044$ & $-0.125$ & -- & $-0.147$ \\
    \midrule
    \multirow{3}{*}{Sem $\uparrow$}
     & PT-prof       & 3.539 & 3.474 & \underline{3.754} & 3.681 & -- & \textbf{3.792} \\
     & Raw-prof      & 3.569 & 3.542 & \textbf{3.779} & 3.708 & -- & \underline{3.765} \\
     & $\Delta$Sem   & $+0.031$ & $+0.068$ & $+0.025$ & $+0.027$ & -- & $-0.027$ \\
    \midrule
    \multirow{3}{*}{Emb $\uparrow$}
     & PT-prof       & 0.394 & 0.400 & \underline{0.414} & 0.412 & -- & \textbf{0.421} \\
     & Raw-prof      & 0.394 & 0.400 & \underline{0.414} & 0.412 & -- & \textbf{0.421} \\
     & $\Delta$Emb   & $0.000$ & $0.000$ & $0.000$ & $0.000$ & -- & $0.000$ \\
    \midrule
    \multicolumn{8}{c}{\textit{Long-Dialogue Macro}} \\
    \midrule
    \multirow{3}{*}{Char $\uparrow$}
     & PT-prof       & 2.326 & 2.454 & \underline{2.989} & 2.904 & 2.894 & \textbf{3.004} \\
     & Raw-prof      & 2.332 & \textbf{3.369} & \underline{3.067} & 3.048 & 2.980 & 2.935 \\
     & $\Delta$Char  & $+0.006$ & $+0.915$ & $+0.078$ & $+0.144$ & $+0.087$ & $-0.069$ \\
    \midrule
    \multirow{3}{*}{Sem $\uparrow$}
     & PT-prof       & 3.323 & 2.774 & \underline{3.474} & 3.332 & 3.363 & \textbf{3.697} \\
     & Raw-prof      & 3.445 & 2.958 & \underline{3.604} & 3.489 & 3.499 & \textbf{3.742} \\
     & $\Delta$Sem   & $+0.123$ & $+0.184$ & $+0.130$ & $+0.157$ & $+0.137$ & $+0.045$ \\
    \midrule
    \multirow{3}{*}{Emb $\uparrow$}
     & PT-prof       & 0.268 & 0.244 & \underline{0.275} & 0.271 & 0.271 & \textbf{0.314} \\
     & Raw-prof      & 0.268 & 0.244 & \underline{0.275} & 0.271 & 0.271 & \textbf{0.314} \\
     & $\Delta$Emb   & $0.000$ & $0.000$ & $0.000$ & $0.000$ & $0.000$ & $0.000$ \\
    \bottomrule
    \end{tabular}
    \end{table}

    \begin{figure}[t!]
      \centering
      \includegraphics[width=\columnwidth]{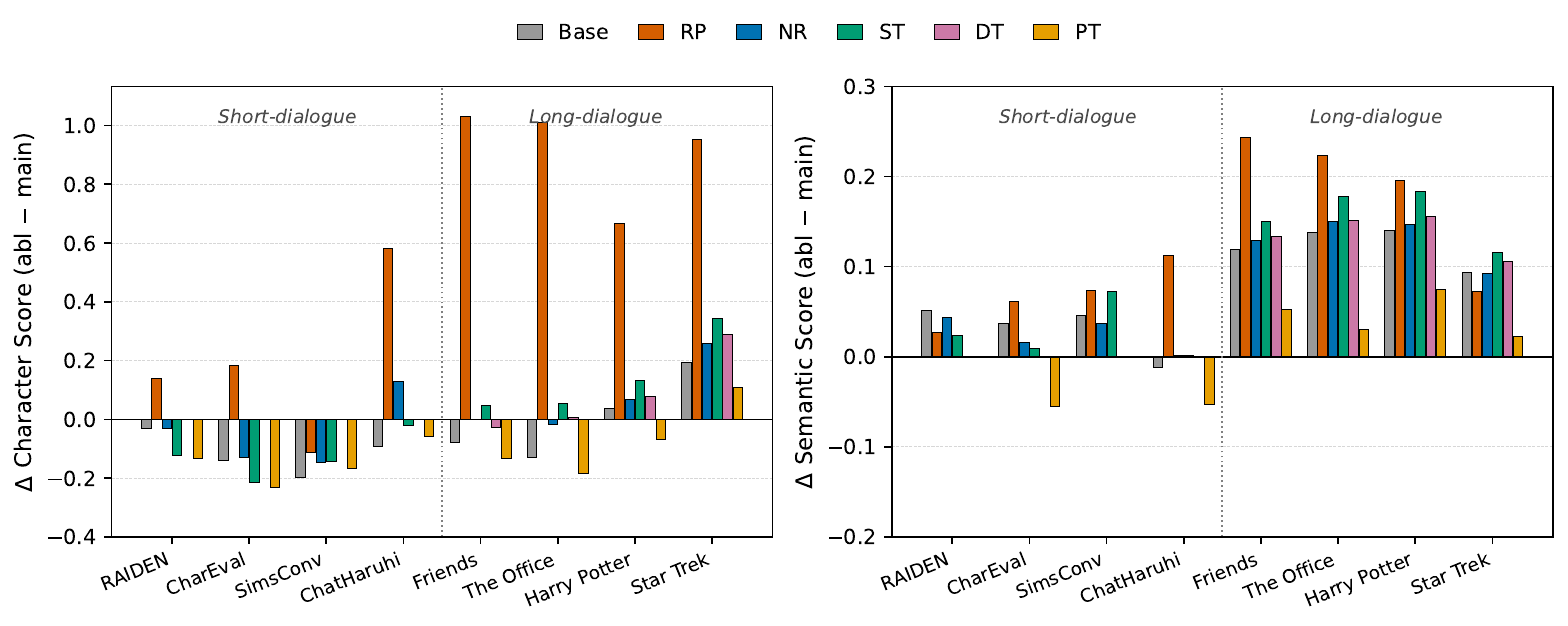}
      \caption{Per-dataset $\Delta$Char (left) and $\Delta$Sem (right) when the judge's persona reference is swapped from PT-prof to Raw-prof (Raw-prof minus PT-prof). RP's Char gain concentrates on long-dialogue corpora and ChatHaruhi; PT stays near zero on both metrics, and the Sem ranking is preserved.}
      \label{fig:judge-ref-ablation}
      \end{figure}

\paragraph{Reference sensitivity.}
Table~\ref{tab:judge-ref-ablation} gives macro-averaged scores under both references. Embedding similarity is unchanged because it does not use the profile reference. Sem varies little on short dialogue ($\pm0.07$) and increases for every method on long dialogue ($+0.04$ to $+0.18$); PT remains first on long-dialogue Sem under both references. Char is more sensitive. RP gains $0.20$ on short dialogue and $0.92$ on long dialogue, whereas NR, ST, and DT move by at most $0.15$. PT decreases by $0.15$ and $0.07$, respectively.

\paragraph{Source of the Char shift.}
The Char rubric measures consistency with the supplied profile text. Raw descriptions in the long-dialogue corpora contain many explicit identity markers, including catchphrases, idioms, and stock mannerisms. RP receives this same description during generation, making its responses more likely to reproduce those markers when the raw description is also used for judging. Accordingly, the largest RP gains occur in the strongly stylized long-dialogue corpora and in ChatHaruhi. The structured profiles distribute character information across fields and place less emphasis on verbatim identity markers, so NR and the tree variants are less sensitive to this reference change.

\paragraph{Relation to Sem and Emb.}
RP retains the lowest macro Sem and Emb among profile-conditioned methods under both reference conditions. Its Char increase therefore reflects closer alignment with the identity markers in the raw description, not a corresponding increase in contextual coherence or reference-response similarity. The effect is also method-specific: RP changes substantially, while the structured variants move only modestly.

\paragraph{Summary.}
The two reference conditions produce the same overall Sem and Emb pattern, while Char responds to the lexical content of the profile reference. Figure~\ref{fig:judge-ref-ablation} shows that this sensitivity is concentrated in RP and in corpora with strongly stylized character descriptions; PT changes only modestly across references.

\subsection{Robustness Across Judge Models}
\label{app:judge-crossmodel}

The judge-model analysis uses GPT-4.1, GLM-5.2, and DeepSeek-V4-Flash with the same rubric, prompt, and decoding settings. Each model scores the same generated responses; embedding scores are shared across conditions.

\begin{figure}[htbp]
  \centering
  \begin{subfigure}[t]{0.49\textwidth}
      \centering
      \includegraphics[width=\linewidth]{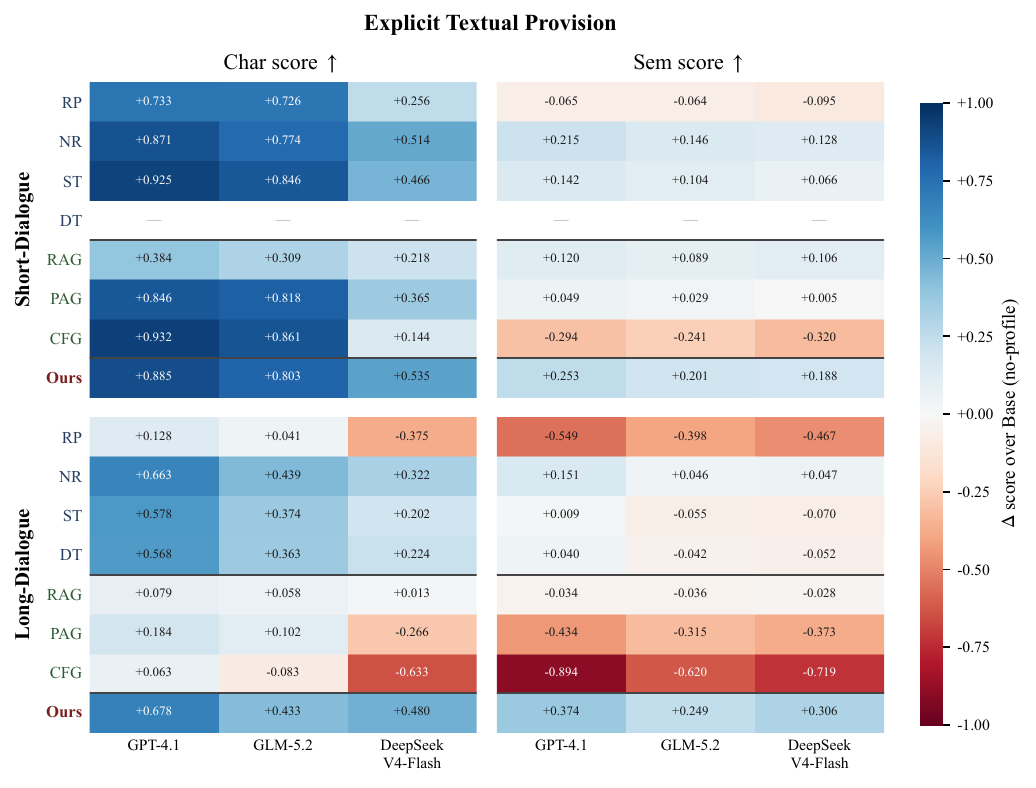}
      \caption{Explicit textual provision.}
      \label{fig:judge-delta-explicit}
  \end{subfigure}
  \hfill
  \raisebox{0.018\textwidth}[0pt][0pt]{%
      \begin{subfigure}[t]{0.49\textwidth}
          \centering
          \includegraphics[width=\linewidth,height=0.72\linewidth]{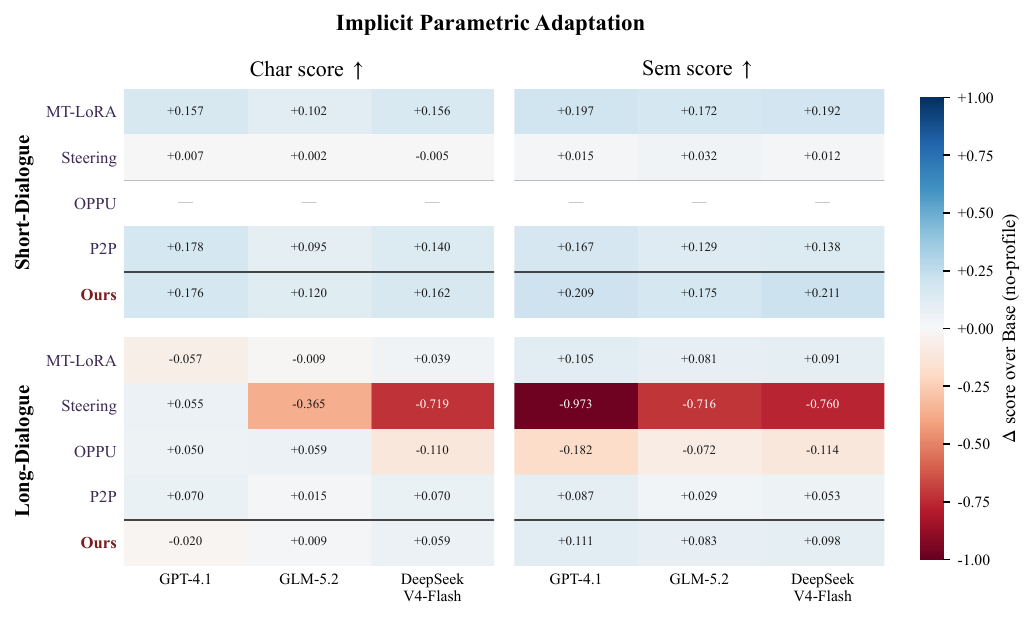}
          \par\vspace*{0.03\linewidth}
          \caption{Implicit parametric adaptation.}
          \label{fig:judge-delta-implicit}
      \end{subfigure}%
  }
  \caption{Character and semantic score change relative to the no-profile Base under three judges (GPT-4.1, GLM-5.2, DeepSeek-V4-Flash), split by horizon. Warmer cells denote larger gains over Base.}
  \label{fig:judge-delta}
\end{figure}

Figure~\ref{fig:judge-delta} (\subref{fig:judge-delta-explicit},~\subref{fig:judge-delta-implicit}) shows each method's change relative to the no-profile Base. Under textual provision, Ours has the largest Sem gain in all six judge--horizon cells, whereas Char leaders vary across judges and horizons. Under parametric adaptation, Ours again leads Sem in all six cells, while its Char gain leads only the short-dialogue cells under GLM-5.2 and DeepSeek-V4-Flash. Thus the Sem conclusion is stable across judges, whereas Char rankings are judge-dependent.

\subsection{Agreement with Human Judgments}
\label{app:human-eval}

Three annotators independently rated a blind sample of 200 Qwen2.5-7B generations spanning all methods and tracks. They used the same 1--5 Character and Semantic scales and saw neither method identities nor automatic scores. Overall is computed post hoc as the arithmetic mean of Char and Sem. Table~\ref{tab:human-eval} reports agreement, judge correlation, and the PT--NR comparison under textual provision.

\begin{table}[htbp]
\centering
\small
\setlength{\tabcolsep}{6pt}
\renewcommand{\arraystretch}{1.15}
\caption{Human evaluation on 200 blinded responses (Qwen2.5-7B), rated by three annotators on 1--5 scales. The last block compares PT with NR on separate $n=10$ prompt subsets.}
\label{tab:human-eval}
\begin{tabularx}{0.56\columnwidth}{@{}>{\raggedright\arraybackslash}Xccc@{}}
\toprule
 & \textbf{Char} & \textbf{Sem} & \textbf{Overall} \\
\midrule
\multicolumn{4}{@{}l}{\textit{Annotator agreement} ($\alpha$)} \\
\quad A/B/C & 0.614 & 0.574 & 0.636 \\
\addlinespace
\multicolumn{4}{@{}l}{\textit{Correlation with judge}} \\
\quad Pearson $r$ & 0.588 & 0.651 & 0.654 \\
\quad Spearman $\rho$ & 0.591 & 0.632 & 0.649 \\
\addlinespace
\multicolumn{4}{@{}l}{\textit{PT vs.\ NR} ($\Delta$)} \\
\quad Prompt track & $+0.067$ & $+0.333$ & $+0.200$ \\
\bottomrule
\end{tabularx}
\end{table}

Annotator agreement ranges from $0.57$ to $0.64$, and human consensus correlates with the judge at $r=0.59$ on Char and $r=0.65$ on Sem and Overall. On the separate $n=10$ PT and NR prompt subsets, the human Overall difference is $+0.20$; this comparison is descriptive. Across 50,232 matched question IDs, GPT-4.1 yields pooled Overall $\Delta=+0.087$ (Wilcoxon signed-rank $p<0.001$).

\FloatBarrier
\section{Backbone Generalization}
\label{app:backbone}

The generator-backbone study covers Qwen3-0.6B, Gemma-4-E4B, Qwen2.5-7B-Instruct, and Qwen3-32B, spanning two model families and more than an order of magnitude in scale. Qwen2.5-7B uses the full test set; the three additional backbones use the same fixed 25\% subsample with identical sample IDs. The extraction pipeline, prompts, decoding configuration, and judge are otherwise fixed. Tables~\ref{tab:backbone-ablation} and~\ref{tab:backbone-external} report horizon-level Char and Sem.

\begin{table}[t!]
\centering
\small
\caption{Cross-backbone internal ablation under explicit textual provision. Qwen2.5-7B uses the full test set; the other backbones use the same fixed 25\% subsample. Values are horizon-level macro-averages. \textbf{Bold} = best, \underline{underline} = second best per row; DT applies to long-dialogue corpora only (--).}
\label{tab:backbone-ablation}
\begin{tabular}{lll cccccc}
\toprule
\textbf{Base LLM} & \textbf{Horizon} & \textbf{Metric} & \textbf{Base} & \textbf{RP} & \textbf{NR} & \textbf{ST} & \textbf{DT} & \textbf{PT (Ours)} \\
\midrule
\multirow{4}{*}{Qwen3-0.6B}
 & \multirow{2}{*}{\shorttag{Short-Dialogue}} & Char $\uparrow$ & 1.642 & \underline{1.968} & \textbf{2.067} & 1.957 & -- & 1.850 \\
 &                        & Sem $\uparrow$  & 2.625 & 2.533 & \textbf{2.803} & 2.718 & -- & \underline{2.744} \\
 & \multirow{2}{*}{\longtag{Long-Dialogue}}  & Char $\uparrow$ & 1.904 & 1.838 & \textbf{2.108} & 1.999 & 1.973 & \underline{2.029} \\
 &                        & Sem $\uparrow$  & 2.861 & 2.336 & \underline{2.916} & 2.799 & 2.763 & \textbf{2.924} \\
\midrule
\multirow{4}{*}{Gemma-4-E4B}
 & \multirow{2}{*}{\shorttag{Short-Dialogue}} & Char $\uparrow$ & 1.949 & 3.001 & 3.057 & \textbf{3.185} & -- & \underline{3.142} \\
 &                        & Sem $\uparrow$  & 3.402 & 3.311 & \textbf{3.651} & 3.482 & -- & \underline{3.619} \\
 & \multirow{2}{*}{\longtag{Long-Dialogue}}  & Char $\uparrow$ & 2.038 & 2.591 & 2.905 & 2.896 & \underline{2.908} & \textbf{3.019} \\
 &                        & Sem $\uparrow$  & 3.256 & 2.941 & \underline{3.456} & 3.304 & 3.316 & \textbf{3.715} \\
\midrule
\multirow{4}{*}{Qwen2.5-7B}
 & \multirow{2}{*}{\shorttag{Short-Dialogue}} & Char $\uparrow$ & 2.143 & 2.876 & 3.014 & \textbf{3.068} & -- & \underline{3.028} \\
 &                        & Sem $\uparrow$  & 3.539 & 3.474 & \underline{3.754} & 3.681 & -- & \textbf{3.792} \\
 & \multirow{2}{*}{\longtag{Long-Dialogue}}  & Char $\uparrow$ & 2.326 & 2.454 & \underline{2.989} & 2.904 & 2.894 & \textbf{3.004} \\
 &                        & Sem $\uparrow$  & 3.323 & 2.774 & \underline{3.474} & 3.332 & 3.363 & \textbf{3.697} \\
\midrule
\multirow{4}{*}{Qwen3-32B}
 & \multirow{2}{*}{\shorttag{Short-Dialogue}} & Char $\uparrow$ & 3.197 & 3.807 & 3.880 & \underline{3.982} & -- & \textbf{3.986} \\
 &                        & Sem $\uparrow$  & 3.983 & 3.890 & \underline{4.062} & 3.993 & -- & \textbf{4.120} \\
 & \multirow{2}{*}{\longtag{Long-Dialogue}}  & Char $\uparrow$ & 2.930 & 3.270 & 3.467 & 3.491 & \underline{3.518} & \textbf{3.685} \\
 &                        & Sem $\uparrow$  & 3.632 & 3.421 & \underline{3.749} & 3.655 & 3.680 & \textbf{4.056} \\
\bottomrule
\end{tabular}
\end{table}

\begin{table}[t!]
\centering
\small
\setlength{\tabcolsep}{9pt}
\caption{Cross-backbone external comparison under explicit textual provision. Qwen2.5-7B uses the full test set; the other backbones use the same fixed 25\% subsample. Values are horizon-level macro-averages. \textbf{Bold} = best, \underline{underline} = second best per row.}
\label{tab:backbone-external}
\begin{tabular}{lll ccccc}
\toprule
\textbf{Base LLM} & \textbf{Horizon} & \textbf{Metric} & \textbf{Base} & \textbf{RAG} & \textbf{PAG} & \textbf{CFG} & \textbf{Ours} \\
\midrule
\multirow{4}{*}{Qwen3-0.6B}
 & \multirow{2}{*}{\shorttag{Short-Dialogue}} & Char $\uparrow$ & 1.642 & 1.681 & \underline{1.970} & \textbf{2.038} & 1.850 \\
 &                        & Sem $\uparrow$  & \underline{2.625} & 2.561 & 2.596 & 2.254 & \textbf{2.744} \\
 & \multirow{2}{*}{\longtag{Long-Dialogue}}  & Char $\uparrow$ & 1.904 & 1.831 & \underline{1.906} & 1.711 & \textbf{2.029} \\
 &                        & Sem $\uparrow$  & \underline{2.861} & 2.511 & 2.481 & 1.917 & \textbf{2.924} \\
\midrule
\multirow{4}{*}{Gemma-4-E4B}
 & \multirow{2}{*}{\shorttag{Short-Dialogue}} & Char $\uparrow$ & 1.949 & 2.315 & 3.049 & \underline{3.130} & \textbf{3.142} \\
 &                        & Sem $\uparrow$  & 3.402 & \underline{3.525} & 3.416 & 3.075 & \textbf{3.619} \\
 & \multirow{2}{*}{\longtag{Long-Dialogue}}  & Char $\uparrow$ & 2.038 & 2.142 & 2.495 & \underline{2.698} & \textbf{3.019} \\
 &                        & Sem $\uparrow$  & 3.256 & \underline{3.270} & 2.911 & 2.774 & \textbf{3.715} \\
\midrule
\multirow{4}{*}{Qwen2.5-7B}
 & \multirow{2}{*}{\shorttag{Short-Dialogue}} & Char $\uparrow$ & 2.143 & 2.527 & 2.989 & \textbf{3.075} & \underline{3.028} \\
 &                        & Sem $\uparrow$  & 3.539 & \underline{3.659} & 3.588 & 3.245 & \textbf{3.792} \\
 & \multirow{2}{*}{\longtag{Long-Dialogue}}  & Char $\uparrow$ & 2.326 & 2.405 & \underline{2.510} & 2.389 & \textbf{3.004} \\
 &                        & Sem $\uparrow$  & \underline{3.323} & 3.289 & 2.889 & 2.429 & \textbf{3.697} \\
\midrule
\multirow{4}{*}{Qwen3-32B}
 & \multirow{2}{*}{\shorttag{Short-Dialogue}} & Char $\uparrow$ & 3.197 & 3.356 & \underline{3.815} & 3.797 & \textbf{3.986} \\
 &                        & Sem $\uparrow$  & 3.983 & \underline{4.004} & 3.910 & 3.622 & \textbf{4.120} \\
 & \multirow{2}{*}{\longtag{Long-Dialogue}}  & Char $\uparrow$ & 2.930 & 2.976 & \underline{3.208} & 3.101 & \textbf{3.685} \\
 &                        & Sem $\uparrow$  & \underline{3.632} & 3.594 & 3.412 & 3.045 & \textbf{4.056} \\
\bottomrule
\end{tabular}
\end{table}

\paragraph{External comparison.}
PHASE-Tree achieves the best long-dialogue Char and Sem on all four backbones and the best short-dialogue Sem on all four. On short-dialogue Char, it ranks first or second except on Qwen3-0.6B. The long-dialogue Sem margin over the strongest competitor is $+0.06$ for Qwen3-0.6B and $+0.37$ to $+0.45$ for the three larger backbones. These estimates support robustness across model families and scales but do not define a monotone capacity trend because the sampling scopes differ.

\paragraph{Internal ablation.}
PT has the best long-dialogue Sem on all four backbones and the best long-dialogue Char on Gemma-4-E4B, Qwen2.5-7B, and Qwen3-32B. On Qwen3-0.6B, NR leads PT on long-dialogue Char ($2.108$ vs.\ $2.029$), while PT remains first on Sem. The PT--NR long-dialogue Char contrast is $-0.079$ at 0.6B, $+0.114$ for Gemma-4-E4B, $+0.015$ at 7B, and $+0.218$ at 32B. These heterogeneous estimates support robustness across model families rather than a scaling curve.

\section{Implicit Parametric Adaptation Results}
\label{app:implicit-results}

Table~\ref{tab:implicit} reports the internal representation ablation for implicit parametric adaptation: the same PHASE-Tree-finetuned hypernetwork generates LoRA adapters conditioned on RP, NR, ST, DT, or PT, with dialogue-only prompts.

    \begin{table}[t!]
    \centering
    \small
    \caption{Implicit parametric adaptation results on eight corpora (mean over random and OOD splits; short/long rows are unweighted macro-averages). \textbf{Bold} = best, \underline{underline} = second best; Base excluded from that competition; DT applies to long-dialogue sets only (--).}
    \label{tab:implicit}
    \begin{tabular}{ll c cccc c}
    \toprule
    \textbf{Dataset} & \textbf{Metric} & \textbf{Base} & \textbf{RP} & \textbf{NR} & \textbf{ST} & \textbf{DT} & \textbf{PT (Ours)} \\
    \midrule
    \multirow{3}{*}{\shorttag{RAIDEN}}
     & Char $\uparrow$ & 2.163 & 2.487 & \textbf{2.514} & \underline{2.513} & -- & 2.510 \\
     & Sem $\uparrow$  & 3.632 & 3.907 & \textbf{3.919} & \underline{3.916} & -- & 3.915 \\
     & Emb $\uparrow$  & 0.444 & 0.504 & \underline{0.506} & \textbf{0.507} & -- & 0.505 \\
    \midrule
    \multirow{3}{*}{\shorttag{CharacterEval}}
     & Char $\uparrow$ & 2.188 & 2.338 & \textbf{2.351} & \underline{2.339} & -- & 2.334 \\
     & Sem $\uparrow$  & 3.382 & 3.549 & \underline{3.550} & \textbf{3.555} & -- & 3.548 \\
     & Emb $\uparrow$  & 0.325 & 0.344 & \underline{0.347} & \textbf{0.348} & -- & 0.345 \\
    \midrule
    \multirow{3}{*}{\shorttag{SimsConv}}
     & Char $\uparrow$ & 2.339 & 2.404 & \underline{2.463} & \textbf{2.508} & -- & 2.455 \\
     & Sem $\uparrow$  & 3.749 & 3.946 & 3.935 & \underline{3.969} & -- & \textbf{3.977} \\
     & Emb $\uparrow$  & 0.439 & \underline{0.518} & 0.510 & 0.517 & -- & \textbf{0.523} \\
    \midrule
    \multirow{3}{*}{\shorttag{ChatHaruhi}}
     & Char $\uparrow$ & 1.880 & \textbf{1.978} & 1.942 & 1.948 & -- & \underline{1.976} \\
     & Sem $\uparrow$  & 3.391 & \underline{3.556} & 3.546 & \textbf{3.564} & -- & 3.551 \\
     & Emb $\uparrow$  & 0.367 & 0.412 & 0.412 & \textbf{0.413} & -- & \underline{0.412} \\
    \midrule
    \multirow{3}{*}{\longtag{Friends}}
     & Char $\uparrow$ & 2.304 & \underline{2.268} & 2.204 & \textbf{2.270} & 2.240 & 2.205 \\
     & Sem $\uparrow$  & 3.303 & \underline{3.422} & 3.418 & \textbf{3.427} & 3.421 & 3.416 \\
     & Emb $\uparrow$  & 0.262 & 0.277 & \textbf{0.278} & 0.277 & 0.278 & \underline{0.278} \\
    \midrule
    \multirow{3}{*}{\longtag{The Office}}
     & Char $\uparrow$ & 2.102 & 2.110 & \textbf{2.113} & \underline{2.112} & 2.070 & 2.112 \\
     & Sem $\uparrow$  & 3.368 & \textbf{3.504} & \underline{3.503} & 3.497 & 3.489 & 3.498 \\
     & Emb $\uparrow$  & 0.254 & 0.267 & 0.267 & 0.267 & \textbf{0.269} & \underline{0.267} \\
    \midrule
    \multirow{3}{*}{\longtag{Harry Potter}}
     & Char $\uparrow$ & 2.342 & \textbf{2.392} & 2.367 & \underline{2.383} & 2.383 & 2.381 \\
     & Sem $\uparrow$  & 3.257 & 3.397 & 3.391 & \underline{3.398} & \textbf{3.402} & 3.396 \\
     & Emb $\uparrow$  & 0.273 & 0.291 & 0.289 & \textbf{0.291} & \underline{0.291} & 0.290 \\
    \midrule
    \multirow{3}{*}{\longtag{Star Trek}}
     & Char $\uparrow$ & 2.557 & \textbf{2.552} & 2.528 & \underline{2.547} & 2.545 & 2.528 \\
     & Sem $\uparrow$  & 3.363 & \textbf{3.434} & 3.415 & \underline{3.432} & 3.428 & 3.426 \\
     & Emb $\uparrow$  & 0.283 & 0.296 & \underline{0.297} & 0.296 & 0.296 & \textbf{0.297} \\
    \midrule
    \multicolumn{8}{c}{\textit{Average Performance}} \\
    \midrule
    \multirow{3}{*}{\shorttag{Short-Dialogue}}
     & Char $\uparrow$ & 2.143 & 2.302 & 2.318 & \textbf{2.327} & -- & \underline{2.319} \\
     & Sem $\uparrow$  & 3.539 & 3.740 & 3.738 & \textbf{3.751} & -- & \underline{3.748} \\
     & Emb $\uparrow$  & 0.394 & 0.445 & 0.444 & \underline{0.446} & -- & \textbf{0.446} \\
    \midrule
    \multirow{3}{*}{\longtag{Long-Dialogue}}
     & Char $\uparrow$ & 2.326 & \textbf{2.330} & 2.303 & \underline{2.328} & 2.309 & 2.306 \\
     & Sem $\uparrow$  & 3.323 & \textbf{3.439} & 3.432 & \underline{3.439} & 3.435 & 3.434 \\
     & Emb $\uparrow$  & 0.268 & 0.283 & 0.283 & 0.283 & \textbf{0.283} & \underline{0.283} \\
    \bottomrule
    \end{tabular}
    \end{table}

\paragraph{Within-ablation analysis.}
Across the conditioned variants, the macro-average spread is only $0.013/0.007$ on Sem and $0.002/{<}0.001$ on Emb for short/long dialogue, indicating that the profile-to-LoRA mapping flattens distinctions among tree variants. ST is the strongest overall internal variant: it leads short-dialogue Char and Sem, is a close second on short-dialogue Emb, and remains within $0.003$ of the best on every long-dialogue metric.

\paragraph{Interpretation.}
Relative to Base, PT changes Char/Sem/Emb by $+0.176/+0.209/+0.053$ on short dialogue and $-0.020/+0.111/+0.015$ on long dialogue. Against external parametric baselines, Ours ranks first in 8 of 24 dataset--metric cells and in the top two in 18, including the highest Sem average on both horizons. The mapping therefore retains a broad conditioning benefit at fixed prompt cost, while the compressed differences among variants identify the profile-to-LoRA encoder as the bottleneck.

\section{Token Cost Analysis}
\label{app:token-cost}

\begin{figure}[!t]
    \centering
    \small
    \setlength{\tabcolsep}{6pt}
    \captionsetup{type=table}
    \caption{Mean token statistics per method, averaged over all datasets and splits within each horizon group. \explicittag{Green} = textual provision; \implicittag{purple} = parametric adaptation. \dag\,CFG performs dual forward passes; its Context and Prompt counts are the sum of the conditioned and unconditioned inputs.}
    \label{tab:token-cost}
    \begin{tabular}{l ccccc ccccc}
    \toprule
    & \multicolumn{5}{c}{\shorttag{Short-Dialogue (Avg)}} & \multicolumn{5}{c}{\longtag{Long-Dialogue (Avg)}} \\
    \cmidrule(lr){2-6}\cmidrule(lr){7-11}
    \textbf{Method} & \textbf{Profile} & \textbf{Context} & \textbf{Prompt} & \textbf{Pred} & \textbf{GT}
                    & \textbf{Profile} & \textbf{Context} & \textbf{Prompt} & \textbf{Pred} & \textbf{GT} \\
    \midrule
    \explicittag{Base}   & 0 & 121 & 204 & 10.5 & 41.2 & 0 & 290 & 372 & 10.3 & 18.0 \\
    \explicittag{RP}     & 416 & 121 & 627 & 17.8 & 41.2 & 275 & 290 & 653 & 13.1 & 18.0 \\
    \explicittag{NR}     & 236 & 121 & 447 & 21.3 & 41.2 & 209 & 290 & 587 & 15.3 & 18.0 \\
    \explicittag{ST}     & 209 & 121 & 421 & 20.0 & 41.2 & 183 & 290 & 561 & 15.0 & 18.0 \\
    \explicittag{DT}     & -- & -- & -- & -- & -- & 482 & 290 & 860 & 15.0 & 18.0 \\
    \explicittag{RAG}    & 0 & 121 & 622 & 16.8 & 41.2 & 0 & 290 & 1347 & 11.9 & 18.0 \\
    \explicittag{PAG}    & 416 & 121 & 1045 & 21.8 & 41.2 & 275 & 290 & 1628 & 13.5 & 18.0 \\
    \explicittag{CFG\dag} & 416 & 242 & 831 & 22.0 & 41.2 & 275 & 580 & 1024 & 13.3 & 18.0 \\
    \explicittag{Ours} & 260 & 121 & 471 & 20.6 & 41.2 & 1358 & 290 & 1736 & 15.5 & 18.0 \\
    \midrule
    \implicittag{MT-LoRA}  & 0 & 121 & 204 & 12.9 & 41.2 & 0 & 290 & 372 & 10.8 & 18.0 \\
    \implicittag{Steering} & 0 & 121 & 204 & 10.5 & 41.2 & 0 & 290 & 372 & 65.2 & 18.0 \\
    \implicittag{OPPU}     & -- & -- & -- & -- & -- & 0 & 290 & 372 & 14.2 & 18.0 \\
    \implicittag{P2P}      & 0 & 121 & 204 & 12.4 & 41.2 & 0 & 290 & 372 & 12.1 & 18.0 \\
    \implicittag{Ours} & 0 & 121 & 204 & 13.2 & 41.2 & 0 & 290 & 372 & 11.0 & 18.0 \\
    \bottomrule
    \end{tabular}

    \vspace{0.6em}
    \captionsetup{type=figure}
    \includegraphics[width=0.92\columnwidth]{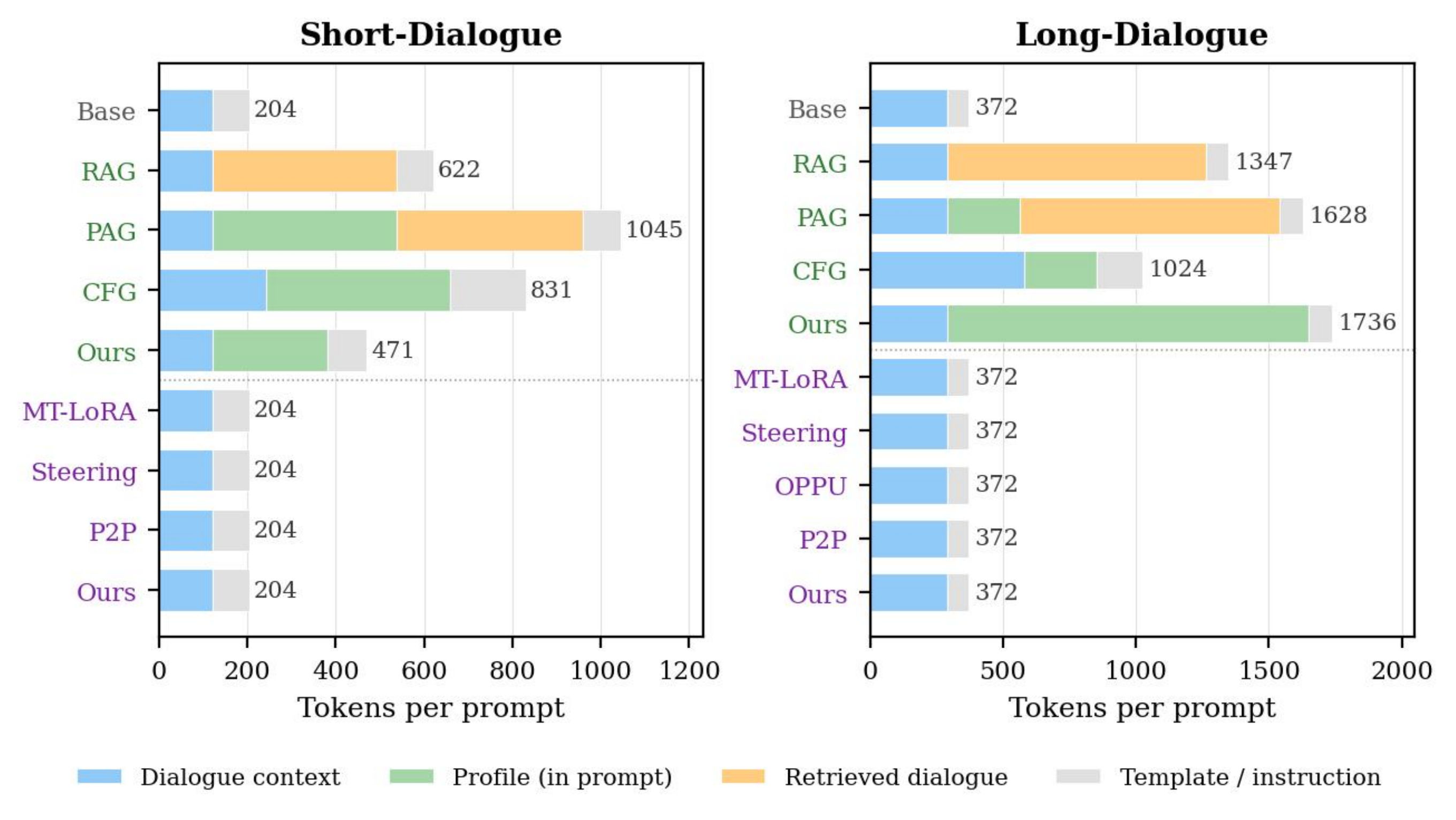}
    \caption{Stacked prompt tokens per method (short/long averages): context (blue), profile (green), retrieved dialogue for RAG/PAG (orange), template/instruction (gray). Right labels: total prompt tokens; dotted line: textual provision vs.\ parametric adaptation. CFG segments are doubled (two forward passes).}
    \label{fig:token-decomposition}
\end{figure}

Table~\ref{tab:token-cost} reports mean token counts per method; within each horizon, values are pooled over its four datasets and both test splits. Profile = character-conditioning tokens in the prompt (0 when encoded in adapters or retrieved as raw dialogue). Context = dialogue-context tokens. Prompt = total LLM input (template, instruction, profile, context). Pred and GT = mean generated and reference response lengths.

\paragraph{Two operating regimes.}
On short dialogue, Ours uses a 471-token prompt, 24--55\% smaller than RP, RAG, PAG, and CFG, while attaining the highest Sem among them. On long dialogue, accumulated state expands the profile to 1358 of 1736 prompt tokens (${\sim}78\%$); this higher input cost yields the strongest Sem and Emb.

\paragraph{Parametric adaptation as a fixed-cost alternative.}
Because character state is absorbed into adapter weights, all parametric methods match the context-only prompt cost (204 short; 372 long), reducing Ours' prompt tokens by ${\sim}57\%$ and ${\sim}79\%$, respectively, relative to textual provision.

Figure~\ref{fig:token-decomposition} decomposes each method's prompt into dialogue context, profile text, retrieved dialogue, and chat-template/instruction overhead. RAG's prompt is dominated by retrieved dialogue (${\sim}418$ tokens short, ${\sim}975$ long); PAG adds the same retrieved dialogue on top of an RP-style profile; CFG roughly doubles context and template overhead because it runs two forward passes per query; Ours under textual provision carries only a profile segment sized by the structured tree; and all parametric-adaptation methods (including Ours under parametric adaptation) reduce to context plus template overhead, with the character state living entirely in the adapter weights.

\end{document}